\documentclass[Afour,sageh,times]{sagej}

\usepackage{moreverb,url}

\usepackage[colorlinks,bookmarksopen,bookmarksnumbered,citecolor=red,urlcolor=red]{hyperref}

\usepackage{orcidlink}

\usepackage{pifont}

\newcommand\BibTeX{{\rmfamily B\kern-.05em \textsc{i\kern-.025em b}\kern-.08em
T\kern-.1667em\lower.7ex\hbox{E}\kern-.125emX}}

\begin{document}

\runninghead{Smith and Wittkopf}

\title{Wukong-Omni: Design, Modeling and Control of a Multi-mode Robot for Air, Land, and Underwater Exploration with All-in-One Propulsion Unit}

\author{Yufan Liu\affilnum{1}, Rixi Yu\affilnum{2}, Junjie Li\affilnum{1}, Yishuai Zeng\affilnum{1}, Zhenting Wen\affilnum{1}, Cheng Li\affilnum{1}, Haifei Zhu\affilnum{2},  Shikang Lian\affilnum{1}, Wei Meng\affilnum{1} and Fumin Zhang\affilnum{3}}
\affiliation{\affilnum{1}School of Automation, Guandong University of Technology, Guangzhou, China\\
\affilnum{2}School of Electromechanical Engineering, Guangdong University of Technology, Guangzhou, China\\
\affilnum{3}Department of Electronic and Computer Engineering, Hong Kong University of Science and Technology, Kowloon, Hong Kong, China,}
\corrauth{Wei Meng, School of Automation, Guandong University of Technology, No. 100 West Outer Ring Road,
Guangzhou Higher Education Mega Center, Panyu District, Guangzhou, 510006, China.}

\email{meng0025@ntu.edu.sg}

\begin{abstract}
In flood disaster rescue scenarios, buildings awaiting search may be partially submerged, preventing aerial robots from entering through lower levels. Conventional robots often need to detour or abandon tasks, posing significant challenges to rescue operations.
To address these limitations, this paper presents design, modeling, control, and performance evaluation of a novel multimode robot named Wukong-Omni. For the first time, this robot enables agile operation across three distinct domains—land, air, and underwater—using a single, unified propulsion system. 
This is made possible through an innovative mechanical design, which enables motor reuse, while its efficiency and peak thrust have been significantly enhanced via simulation and tank-based optimization.
Experimental results reveal that the proposed structure achieves a 100\% improvement in propulsion efficiency and a 150\% increase in maximum thrust compared to direct installation methods, while also enhancing operational stability. Compared with existing cross-domain propulsion units, Wukong-Omni's unit demonstrates significantly higher efficiency and broader environmental locomotion ability.
We have designed an appropriate body structure, analyzed dynamic models for various operational domains, and proposed a unified cross-domain control framework. Comprehensive experiments are conducted individually in three domain, validating the superior performance. Furthermore, outdoor experiments in unstructured environments are carried out, demonstrating Wukong-Omni’s robustness, adaptability, and practical applicability in real-world scenarios.
\end{abstract}


\maketitle

\section{1 Introduction}
\begin{table*}[t]
  \caption{\label{tab:intro}Comparison of Structure and Performance of Representative prototypes}
  \centering
  \resizebox{\textwidth}{!}
  {
  \begin{tabular}{c c c c c c c}
  \hline
  References&Type&Land ability&Water ability&Propulsion number&Air endurance&Water hover endurance\\
  \hline  
  \cite{chaikalis2023mechatronic}&Coaxial tricopter&Active differential&Surface&6&18 mins&None\\  
  \cite{takahashi2015all}&Quadrotor&Passive rolling&Surface&4&None&None\\
  \cite{guo2019system}&Coaxial Quadrotor&Passive rolling&Underwater&8&None&None\\
  Proposed&\textbf{Tilt Quadrotor}&\textbf{Active ackermann}&\textbf{Underwater vector}&\textbf{4}&5 mins&1-2h\\
  \hline
  \end{tabular}
   }
   \label{prototypetable}
  \vspace{-4mm} 
\end{table*}

\begin{figure}[t]
  \centering
  \includegraphics[width=\columnwidth]{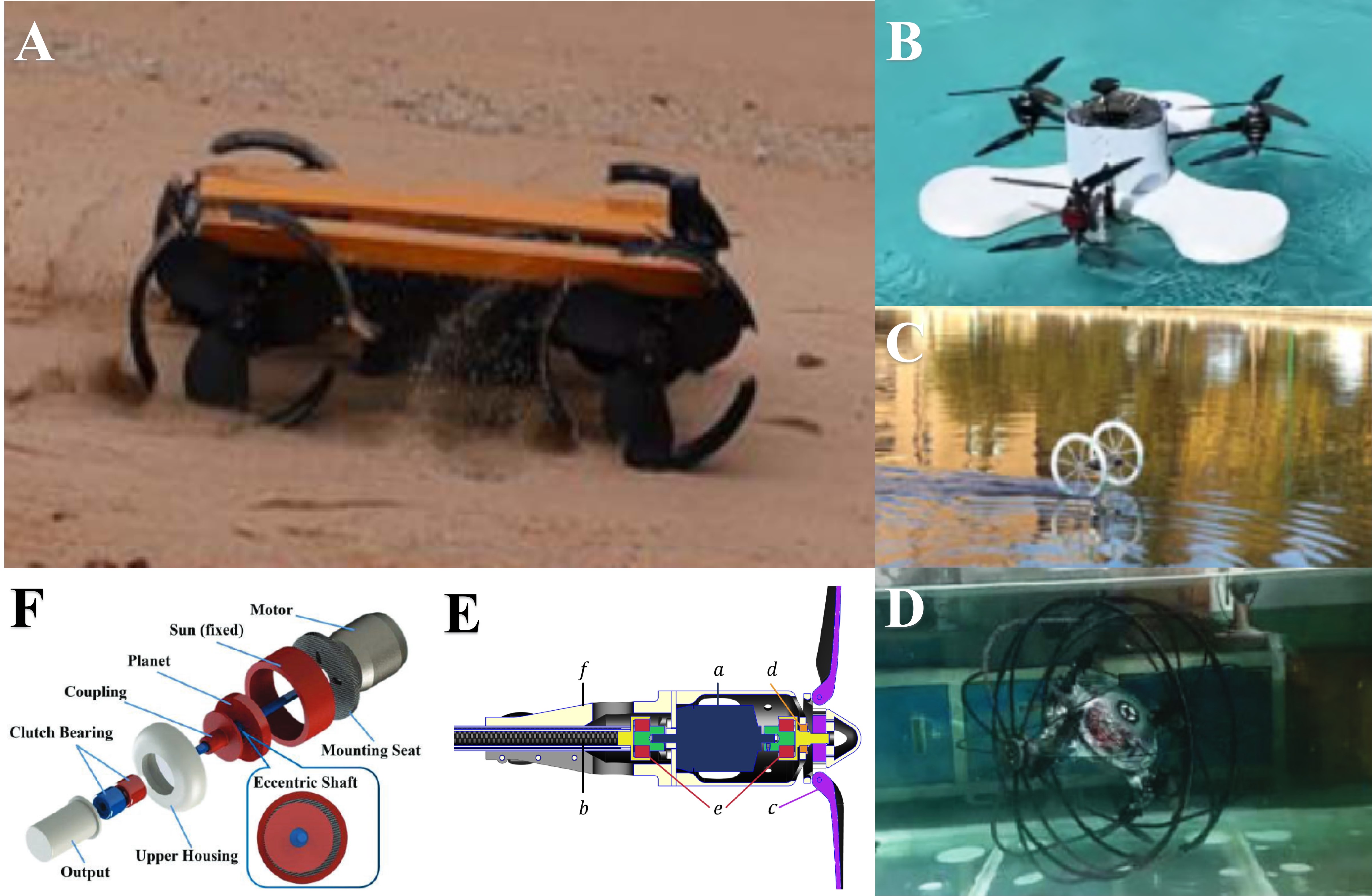}
    \caption{Several representative cross-domain and motor reuse structure prototypes.
    A. SHOALBOT wheel-propeller prototype (\cite{ma2022design}). 
    B. Coaxial tricopter based prototype (\cite{chaikalis2023mechatronic}). 
    C. Two-wheeled based prototype (\cite{takahashi2015all}). 
    D. Caged based prototype (\cite{guo2019system}).
    E. Dipper's motor reuse structure prototype (\cite{rockenbauer2021dipper}).
    F. TJ-FlyingFish's gearbox structure prototype (\cite{liu2023tj}).} 
  \label{intro_pic}
  \vspace{-6mm} 
\end{figure}

With the development of multimode robotics, these systems have found growing applications in tasks carried out in inaccessible environments. They have been employed in data collection missions (\cite{liu2024wukong}), biological and geological sampling  (\cite{2025SailMAV,2025Nezha-Hob,2024WATundICE,2024AASSobser,2019iceob,2024icesf}), offshore facility inspection (\cite{ma2022design,2025WheelPulp}), reconnaissance and disaster rescue operations (\cite{zhang2022autonomous,M42023,2024flood}), and environmental conservation efforts (\cite{cruz2008mares}). These diverse application scenarios are made possible by the ability to locomotion in different domain.

Fundamentally, the significant differences in physical properties among aerial, underwater, and terrestrial environments pose substantial challenges for the design and operation of cross-domain robots. For instance, water is 772.4 times denser and 56.42 times more viscous than air. A key challenge lies in the incompatibility of propulsion systems across these different domains. The distinct requirements of three domains for the size and parameters of propulsion units often necessitate design compromises, which can lead to reduced efficiency in at least one domain (\cite{tan2017efficient}). Due to the difference in fluid densities, propulsion systems must provide different torque outputs to be effective in air and water, while terrestrial locomotion demands rough surfaces to generate sufficient friction. 
Similarly, structural compatibility is limited. For example, while aerodynamic drag can often be neglected in low-speed aerial and terrestrial motion, underwater operation requires streamlined housings and minimized volume and frontal area to reduce hydrodynamic resistance (\cite{yao2023review}). Consequently, cross-domain robots must adopt specialized propulsion configurations and carefully optimized structural designs.

Recently, a number of prototypes for cross-domain robots capable of operating in two distinct environments have been developed, such as air-water or land-water. However, research on robots capable of operating across water, air, and land remains limited, and the development of such prototypes is still in its early stages. 
For instance, some studies use the lightweight inflatable wheels mounted on both sides of a quadrotor to provide buoyancy and support (\cite{takahashi2015all}), but such designs are incapable of underwater operation. 
In the reference (\cite{chaikalis2023mechatronic}), the authors propose to use three separate propulsion systems for aerial, underwater, and ground locomotion. 

In (\cite{guo2019system}), a different design features a caged quadrotor with an octocopter configuration, relying on aerial propellers for underwater propulsion and using the cage structure for ground support. Although this approach avoids significant weight increase, using aerial propellers underwater results in low efficiency, and may damage the Electronic Speed Controller (ESC). 
Another study introduces a screw-drive mechanism added to a quadrotor platform for mobility on land and water surfaces (\cite{canelon2021design}), which results in adding weight to the quadrotor. 
In summary, these prototypes highlight a key challenge: the existing prototypes tend to adopt multiple sets of propulsion systems, which often reduce overall efficiency. Additionally, using mismatched propulsion mechanisms in inappropriate domain further degrades task performance and success rates. Moreover, the mobility of these prototypes is restricted to the water surface, which lacks the capability for submersion. A brief review of these prototypes is shown in Fig. \ref{intro_pic}.D-F. and TABLE \ref{tab:intro}.

To explore these issues, some research groups have begun to explore propulsion reuse technologies. In (\cite{tan2017efficient}), the authors develop a prototype unit composed of one-way bearing, aerial propeller and a gearbox, and conducted performance evaluations. Building upon this, TJ-FlyingFish (\cite{liu2023tj}) integrated the unit into a quadrotor platform. However, both of them overlook the issue of low propulsion efficiency when using aerial propellers underwater. 
SHOALBOT (\cite{ma2022design}) uses wheel-propeller hybrid structures and enhances ground and underwater mobility through optimization. Additionally, Dipper (\cite{rockenbauer2021dipper}) employes one-way bearings to allow a fixed-wing motor to drive both aerial and underwater propellers, achieving promising performance in both aerial and subaqueous environments. These prototypes are shown in Fig. \ref{intro_pic}.A-C. These existing reuse techniques have demonstrated the feasibility of sharing propulsion in two domains. However, a solution capable of effective propulsion across three domains, i.e., air, water, and land, has yet to be realized.

In this paper, we will first develop a multi-mode robot which has the work capability in three domains. The key features of our design is that we propose a novel all-in-one propulsion unit, along with the system design and dynamic modeling of the developed cross-domain robot. Furthermore, we propose an integrated control framework tailored for cross-domain robotic operations. The highlighted performance of the robot will be validated through both simulation and real-world experiments. The main contributions of this work are as follows:

1) A novel all-in-one propulsion solution: Inspired by the wheel-propeller concept and the use of one-way bearings with gear reduction mechanisms, we propose a design capable of operating across three environments. Through comparative experiments and structural constraint-based design of the novel module, we analyze the mutual influence of operations in air, water, and on land.

2) Design of a new cross-domain prototype: Wukong-Omni is developed featuring a tiltable body design that takes into account waterproofing, maneuverability, and structural requirements for seamless transitions across aerial, aquatic, and terrestrial environments.

3) Integrated control framework for cross-domain robots: Through the multimode dynamic model, an integrated control framework is proposed to schedule different modal controllers for the switching control of cross domain robots.

4) Verification of performance through field testing: Conducted extensive field experiments to evaluate the cross-domain mobility and control performance of the Wukong-Omni robot, demonstrating the effectiveness of the proposed unit and validating the integrated control strategy.

The remainder of this paper is organized as follows: Section II introduces the multimode operation and optimization of the all-in-one propulsion unit. Section III presents the prototype design. Section IV describes the robot modeling. Section V outlines the control strategy. Section VI reports experimental results, and Section VII concludes the paper.
\section{2 ALL-IN-ONE PROPULSION UNIT}
\subsection{2.1 Problem Statement}

\begin{figure*}[!t]
  \centering
  \includegraphics[width=\textwidth]{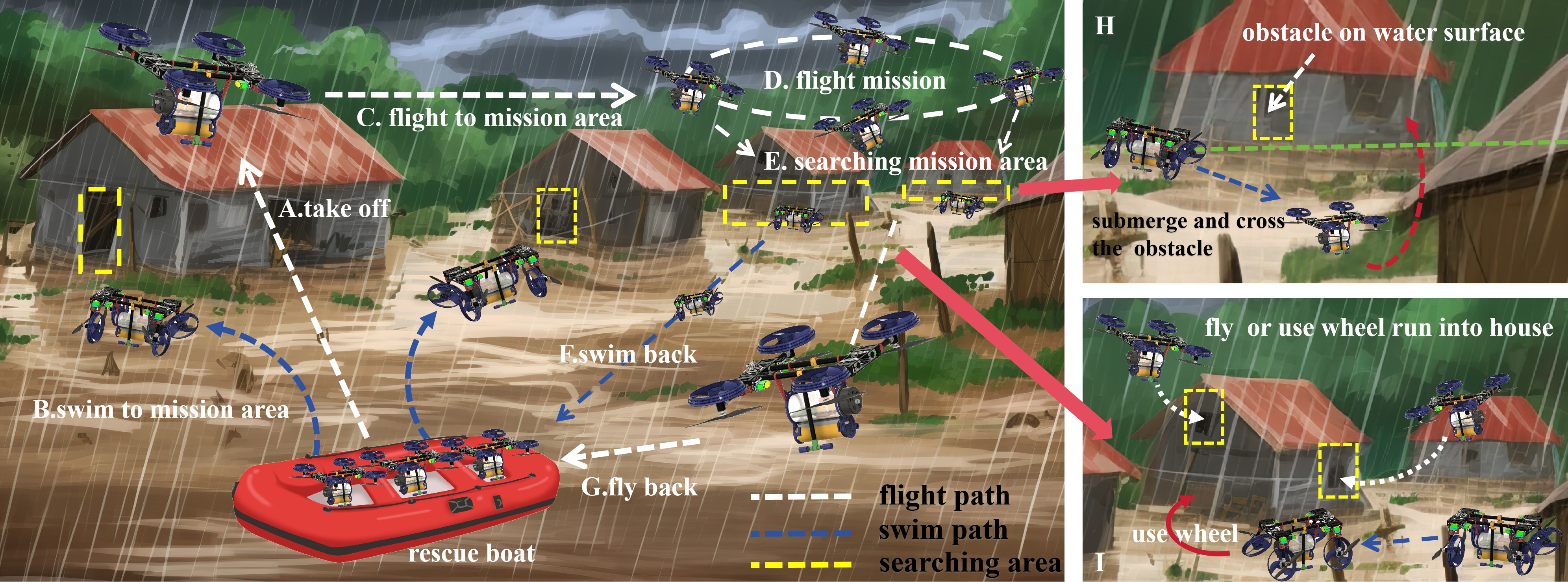}
  \caption{Cross-Domain Operation Scenarios in Flood Rescue Missions.
    \textbf{A}. Taking off from the rescue boat.
    \textbf{B}. Swimming to the operation area.
    \textbf{C}. Flying to the operation area.
    \textbf{D}. Searching for accessible entry points at the operation area.
    \textbf{E}. Performing search and rescue tasks.
    \textbf{F}. Swimming back to the rescue boat via the water surface.
    \textbf{G}. Flying back to the rescue boat.
    \textbf{H}. Diving to access submerged entrances.
    \textbf{I}. Entering through open entrances via flight or rolling.}
  \label{modulebg}
  \vspace{-3mm}
\end{figure*}
\begin{figure*}[!t]
\centering
\includegraphics[width=\textwidth]{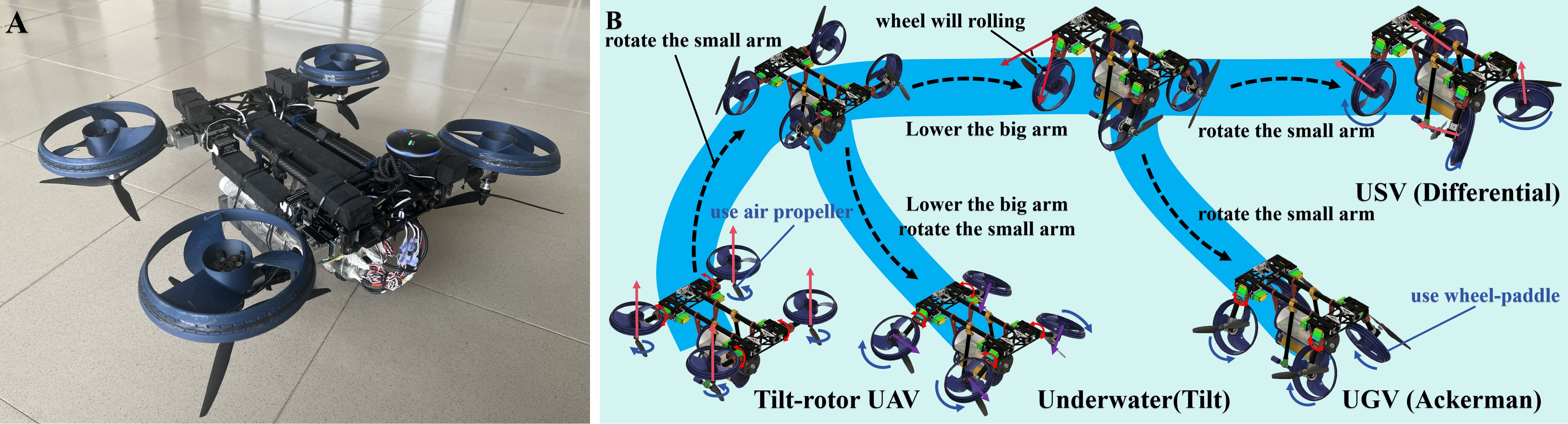}
\caption{\textbf{A}.The prototype of Wukong-Omni. 
           \textbf{B}.The diagram of Wukong-Omni's various modes and the transitions.}
\label{modulestaus}
\vspace{-4mm}
\end{figure*}

In Fig. \ref{modulebg}, one of the potential application scenarios of three-domain tasking robots is presented. As shown in the figure, in the flood rescue missions, the robot can quickly reach the target area using either the aerial or water surface mode to scan for accessible entrances. When open entrances are present, the robot can enter by flying or rolling in land mode (see Fig. \ref{modulebg}.I); if the entrances are submerged, the robot can switch to underwater mode and submerge to explore interior cavities (see Fig. \ref{modulebg}.H). 

\begin{table}[!t]
  \caption{All-in-One Propulsion Unit's Gearbox Param Selection}
  \centering
  \resizebox{\columnwidth}{!}{
  \begin{tabular}{c c c c c}
  \hline
  Parameter & Gears & G.width& Shift coff& Quality\\
  (unit) & (z) &(mm)(b)& -(x)& ISO1328:2013(A)\\
  \hline
  Sun(1)&12&9.6&0.488&Q6\\
  Planetary(1)&30&4.0&-0.488&Q6\\
  Internal gear(2)&78&5.0&-0.195&Q6\\
  Planetary(2)&15&14.4&0.195&Q6\\
  \hline
  \end{tabular}
  }
  \label{gearparam}
  \vspace{-6 mm} 
\end{table}

To handle the tasking scenarios addressed above, in this work, we develop a novel prototype, Wukong-Omni, which supports multiple operation modes to adapt to diverse environments: aerial tilting flight mode, underwater tilting mode, surface differential drive mode, and land Ackermann steering mode. As shown in Fig. \ref{modulebg}, these multimode capabilities allow the robot to flexibly respond to complex scenarios. 

Wukong-Omni's multimode capability is enabled by its reconfigurable body and all-in-one propulsion Unit. This unit integrates gearbox for enhanced torque in water and land modes, wheel-propellers with foot-like supports for terrain adaptation, and bearing-mounted aerial propellers for flight. Designing such a unit requires careful consideration of structural layout, hydrodynamic optimization, and aerodynamic matching for aerial performance. The following sections outline the process of addressing key challenges and the corresponding solutions.

\subsection{2.2 Unit Mechanical Design}
To enable the reuse of both the aerial propeller and the wheel–propeller, we adopt a simple but effective design that employs a dual-output-shaft motor with one-way bearings mounted on both ends, as illustrated in Fig. \ref{Unitpic}.A. When the motor rotates in one direction (e.g. during land or underwater mode, shown in Fig. \ref{Unitpic}.C), it drives the wheel–propeller unit while the aerial propeller remains disengaged. Conversely, when the motor rotates in the opposite direction, the aerial propeller is engaged while the wheel–propeller unit is disengaged, as depicted in Fig. \ref{Unitpic}.B.

To match the aerial motor's torque and speed in underwater mode, planetary gearbox is set on the wheel–propeller side. It features a face-mounted output, external ring gear transmission, and a compact four-point contact bearing on the planet carrier for improved shock resistance and reduced weight.

\textbf{\begin{figure*}[!t]
  \centering
  \includegraphics[width=\textwidth]{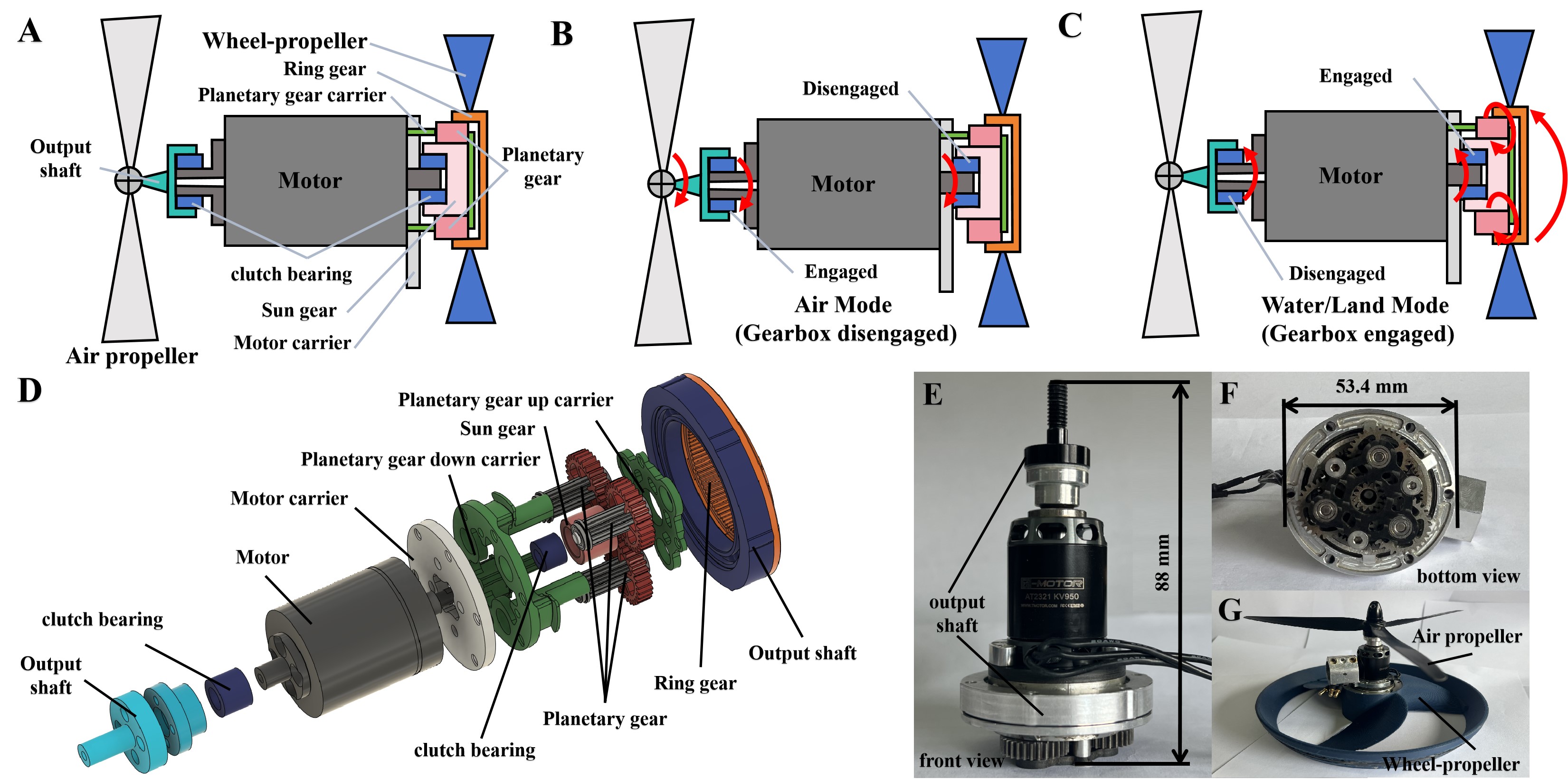}
  \caption{All-in-One Propulsion Unit operation schematic.
  \textbf{A}. The Unit structure composition.
  \textbf{B}. The aerial mode.
  \textbf{C}. The aquatic and land mode. 
  \textbf{D}. The exploded view of the developed unit. 
  \textbf{E}. The front view and size of the unit.
  \textbf{F}. The gearbox structure details of the unit.
  \textbf{G}. The overall structure of the unit.}
  \label{Unitpic}
  \vspace{-6 mm} 
\end{figure*}}

The first stage uses module 0.6 gears (12-tooth sun, 30-tooth planets), while the second stage uses module 0.4 gears (15-tooth planets, 78-tooth ring). Positive profile shift on the planets and negative on the ring ensure proper clearance and meshing. These parameters enable the compact gearbox to deliver high torque in limited space, making it ideal for size and weight constrained multimode robotic systems. The structure and params are shown in Fig. \ref{Unitpic}.D and Table \ref{gearparam}.

In this work,  propeller analysis and design is done by using QPROP (\cite{drela2007qprop}). The relationship between torque $Q$ and rotation speed $\omega$ can be model as 
\begin{equation}
  \begin{aligned}
    Q = [(V-\frac{{\omega}}{K_{vg}} ) \frac{1}R - i_{0}] \frac{1} {K_{vg}},
  \end{aligned}
\end{equation} 
where $V$ is the input voltage, $i_0$ is the no-load current, $R$ is the motor resistance, and $K_{vg}$ is the speed constant (rpm/v) after gearbox reduction. A gearbox with a gear ratio of $G_r\!:\!1$ reduces the motor speed constant $K_v$ as follows:
\begin{equation}
  \begin{aligned}
   K_{vg} = \frac{K_v}{G_{r}}, 
  \end{aligned}
\end{equation} 
and the power loss of the gearbox is modeled as an increase in the zero-load current, as follows
\begin{equation}
  \begin{aligned}
   I_0= I_0 + I (1 - \eta_G), 
  \end{aligned}
\end{equation} 
where $G_r$ represent the gearbox reduce ratios, $K_{v}$ is the rpm/v, $I_0$ is the no-load current, $I$ is the expected operating current, and the $\eta_G$ is the expected gearbox efficiency. The efficiency of gearbox and motor is defined as follows
\begin{equation}
  \begin{aligned}
    \eta_M = \frac{Q {\omega}} {VI}.
  \end{aligned}
\end{equation} 
We use the specific thrust to estimate the all module efficiency and the propeller efficiency, which can be defined as follows
\begin{equation}
  \begin{aligned}
    \eta_f = \frac{T} {VI}, \quad \eta_p = \frac{T} {Q\omega}
  \end{aligned},
\end{equation} 
where $T$ is the thrust. By inputting the selected motor parameters and commonly used underwater propeller geometries into QPROP, we can obtain a preliminary estimate of the gear reduction ratio. The detailed calculation process follows the methodology have also been outlined in (\cite{tan2017efficient}). In addition, practical design constraints considered, including the response time of the wheel–propeller and the mechanical requirements for traversing ground obstacles. Based on these factors, the gear ratio was finalized as 1:13.

Based on the selected motor characteristics, we further determine the approximate dimensions of both underwater and aerial propellers. The detailed process of optimal parameter selection will be discussed in the following section. The key specifications of the unit are summarized in Table \ref{Unitparam}.

\subsection{2.3 Wheel-Propeller Structure Optimization}
\begin{table}[!t]
\caption{Wheel-Propeller Initial Param Selection}
\centering
\resizebox{0.7\columnwidth}{!}{
  \begin{tabular}{c c c}
  \hline
  Parameter & unit & Value \\
  \hline
  Diameter&mm&180\\
  Pitch ratio&-&0.71\\
  Disk ratio&-&0.3056\\
  Blade-number&-&3\\
  Meanline type&-&NACA a=0.8\\
  Thickness type&-&NACA 66 (DTRC)	\\
  Material&-&PLA\\
  Wheel-propeller weight&g&86g	\\
  \hline
  \end{tabular}
}
\label{Unitparam}
\vspace{-5 mm} 
\end{table}

\textbf{1) Selection of the Initial Propeller Radius:}
We adopt commonly used propeller types, with emphasis on selecting suitable underwater propeller dimensions. To prevent ground interference during land-mode operation, the underwater propeller is designed with a slightly smaller radius than the aerial propeller. Blade geometries are parametrically generated using OpenProp (\cite{epps2009openprop}), targeting maximum thrust and efficiency. Design parameters are summarized in Table \ref{Unitparam}.

\begin{figure*}[!t]
  \centering
  \includegraphics[width=\textwidth]{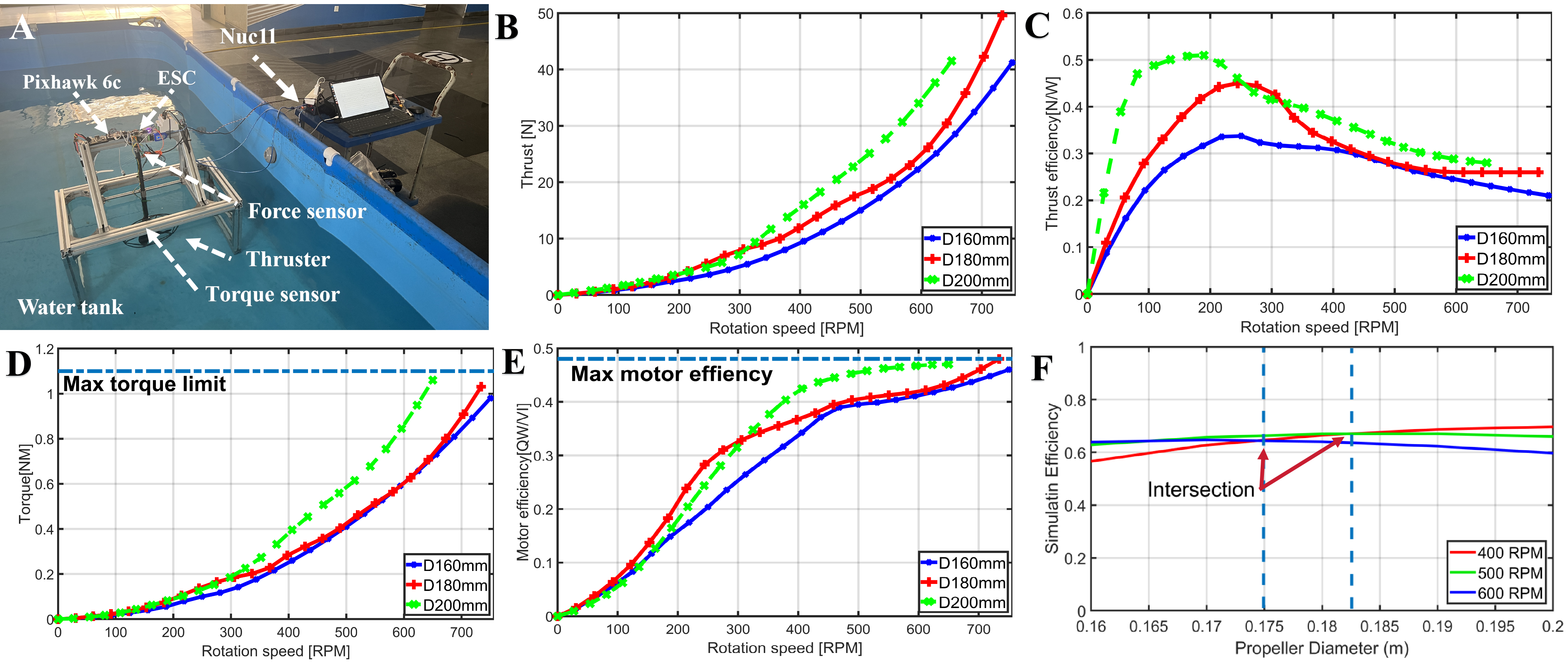}
  \caption{Performance evaluation of propellers in (160 mm, 180 mm, 200 mm) diameters.
    \textbf{A}. The Schematic of the test bench and water tank.
    \textbf{B}. The thrust-rpm curves. 
    \textbf{C}. The efficiency-rpm curves. 
    \textbf{D}. The torque-rpm curves. 
    \textbf{E}. The motor–gearbox efficiency-rpm curves. 
    \textbf{F}. The simulated efficiency-diameters curves using OpenProp.}
  \label{expradius}
  \vspace{-5 mm} 
\end{figure*}

\begin{table}[t]
  \caption{Test bench Component Selection}
  \centering
  \resizebox{0.9\columnwidth}{!}{
  \begin{tabular}{c c}
    \hline
  Equipment & Type \\
  \hline
  Force sensor&RUILIDE RDF-TM20\\
  Torque sensor&RUILIDE RDN-97B\\
  Encoder&blheli dshot Feedback\\
  Driver&Tmotor velox-V45A-V2\\
  Distribution Board& Holybro PM07\\
  Controller&Pixhawk 4 (custom high rate feedback)\\
  Water channel&Length-400 cm, Width-211 cm, Depth-81 cm\\
 \hline
 \end{tabular}
 }
 \label{CSTTABEL}
 \vspace{-2 mm} 
\end{table}

\begin{table}[t]
\caption{Performance Comparison of Different Propeller Sizes in Thrust and Efficiency}
\centering
\resizebox{\columnwidth}{!}{
\begin{tabular}{cccc}
\hline
Propeller size&Max thrust&Max efficiency&Max unit efficiency\\
(mm)&(N)&(N/W)&(Q$\omega$/UI)\\
\hline
D160mm&41.2&0.337&0.46 \\
\textbf{D180mm}&\textbf{49.6}&0.449&\textbf{0.488} \\
D200mm&41.5&\textbf{0.507}&0.47 \\
\hline
\end{tabular}
}
 \vspace{-6 mm} 
\end{table}

Three candidate diameters candidate—160 mm, 180 mm, and 200 mm—were evaluated. OpenProp simulation results under varying speeds are shown in Fig. \ref{expradius}.F. At 500 r/min, the 180 mm propeller exhibited peak efficiency, prompting further analysis centered on this configuration.

To validate these results, we built a test bench (Fig. \ref{expradius}.A), with hardware listed in Table \ref{CSTTABEL}. Thrust was measured via a force sensor on a lever-type balance (kept dry), and torque was captured using a waterproofed sensor fixed underwater at 50Hz.

Test results are shown in Fig. \ref{expradius}.D. Due to the one-way bearing’s mechanical limit, the allowable torque was capped at 1.1 Nm, and tests were confined within this safe range.

The results show that at the same rotational speed, thrust, maximum efficiency, and required torque increase with the diameter of the propeller. However, larger diameters shift the peak efficiency to lower speeds. Between 500–700 r/min, efficiencies for all three configurations converge, with differences under 0.05.

\begin{figure}[t]
  \centering
  \includegraphics[width=0.95\columnwidth]{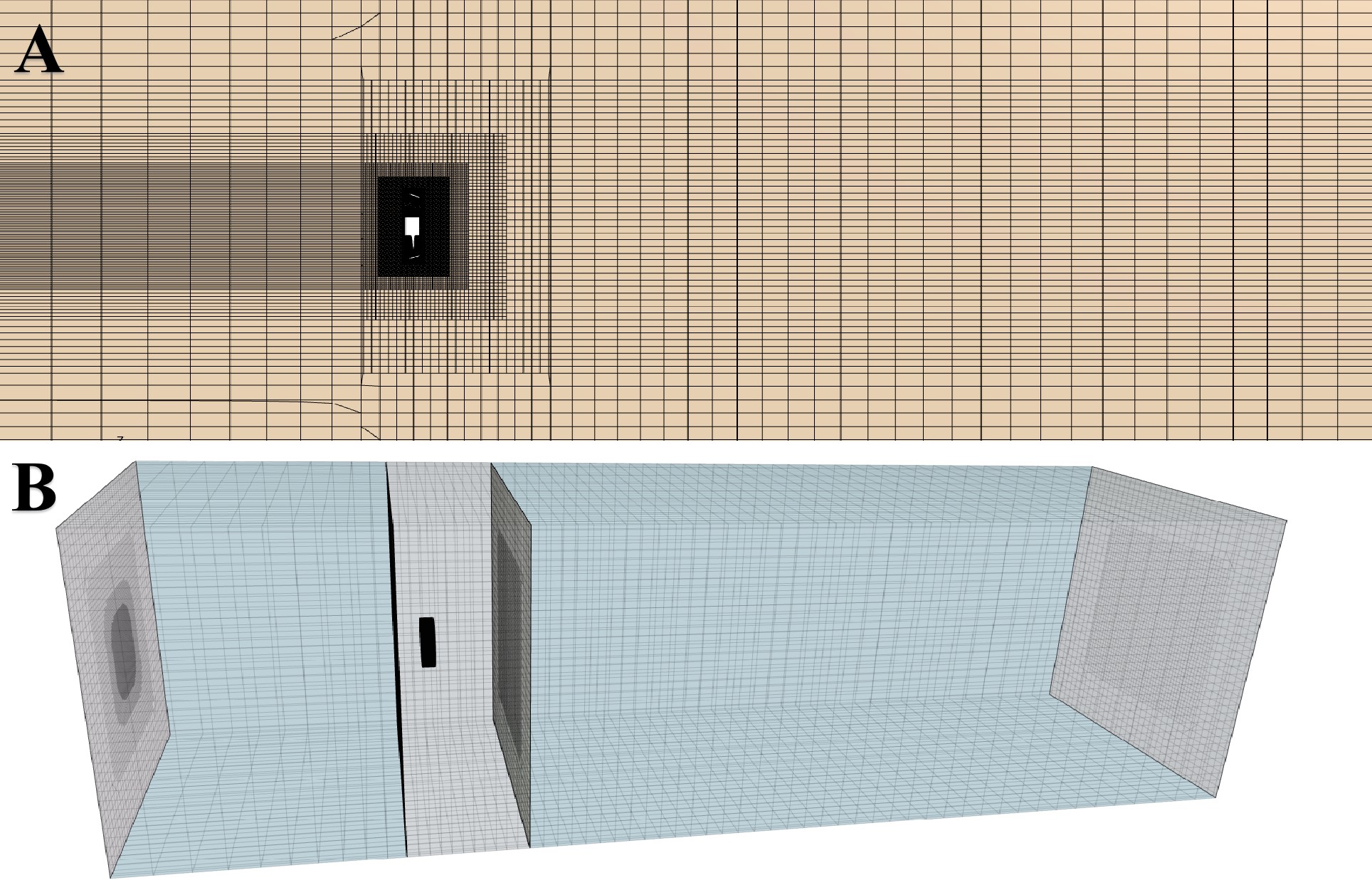}
  \caption{The cross-sectional and 3D views of the mesh during simulation.}
  \vspace{-4 mm} 
  \label{wanges}
\end{figure}

While the 200 mm propeller offers the highest efficiency, it also requires the highest torque. This would necessitate lower speeds to remain within safe limits, reducing thrust. The 180 mm propeller, on the contrary, offers a favorable balance—delivering the highest thrust with moderate torque demands—making it the most suitable for underwater operation.

\begin{figure*}[!t]
  \centering
  \includegraphics[width=\textwidth]{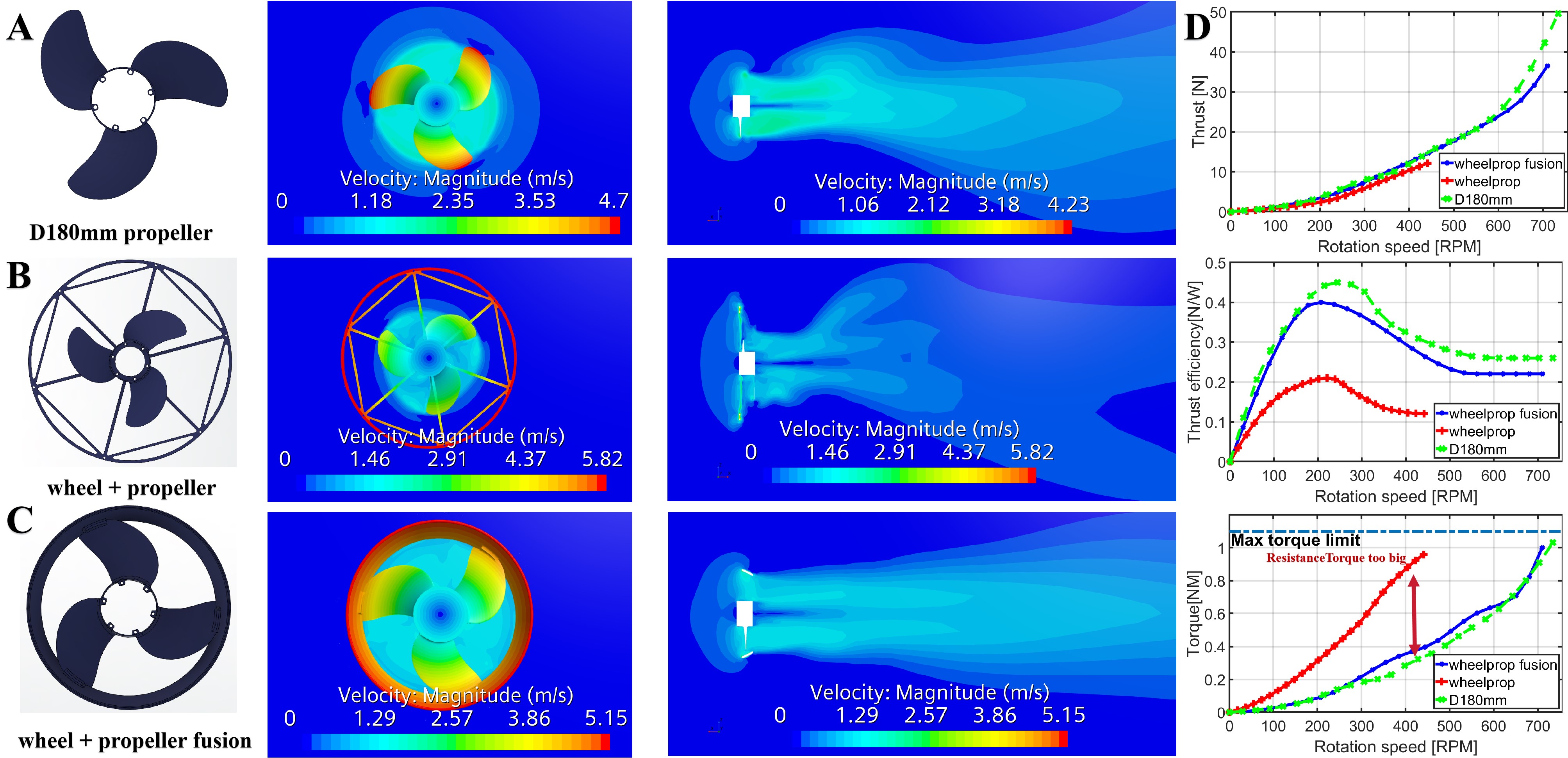}
  \caption{Performance evaluation of propellers in different structure (D180mm, wheel+propeller, wheel-propeller fusion)}
  \vspace{-4 mm} 
  \label{wheeltypesimu}
\end{figure*}

Finally, motor efficiency, as shown in Fig. \ref{expradius}.E, remains consistent around 0.48 across all cases. This indicates that it is largely independent of propeller size and dictated by the motor and gearbox characteristics.

\textbf{2) Selection of the Optimal Propeller Shape:}
After defining the baseline propeller parameters, we investigated how to integrate the wheel structure with the propeller and evaluated its hydrodynamic impact. Initially, slender carbon fiber spokes were used to form a rim enclosing the aerial propeller. This design aimed to prevent ground contact in land mode and provide collision protection, while minimizing cross-sectional area to reduce drag.

As shown in Fig. \ref{wheeltypesimu}.D, the initial configuration generated measurable thrust in water tank tests. However, performance dropped significantly: maximum thrust was reduced by 75.6\% and peak efficiency by 53.4\%. At 450 RPM, torque reached 1 Nm, nearly four times that of the pure propeller, indicating substantial hydrodynamic resistance.

Due to the absence of analytical models for wheel–propeller systems, a CFD simulation environment was developed using Simcenter Star-CCM+ 2402.0001 (19.02.012). A moving reference frame was applied to the propeller-leg structure within a larger stationary domain (Fig. \ref{wanges}.AB), with inflow and outflow regions set to 1000 mm and 3000 mm, and a cross-sectional width seven times the propeller diameter. Trimmed-cell meshing was adopted, using a 5 mm base size in non-critical areas, refined to 25\% on the propeller surface and 5\% around the edges and wheel rim. All simulations used consistent mesh settings to ensure comparability. As shown in Table \ref{simuexperror}, the simulation results closely matched experimental data, with errors ranging from 2.9\% to 15.64\%, confirming that the simulations provide a reliable reference for analyzing hydrodynamic performance.

\begin{figure}[t]
\hspace*{-0.5cm} 
  \centering
  \includegraphics[width=\columnwidth]{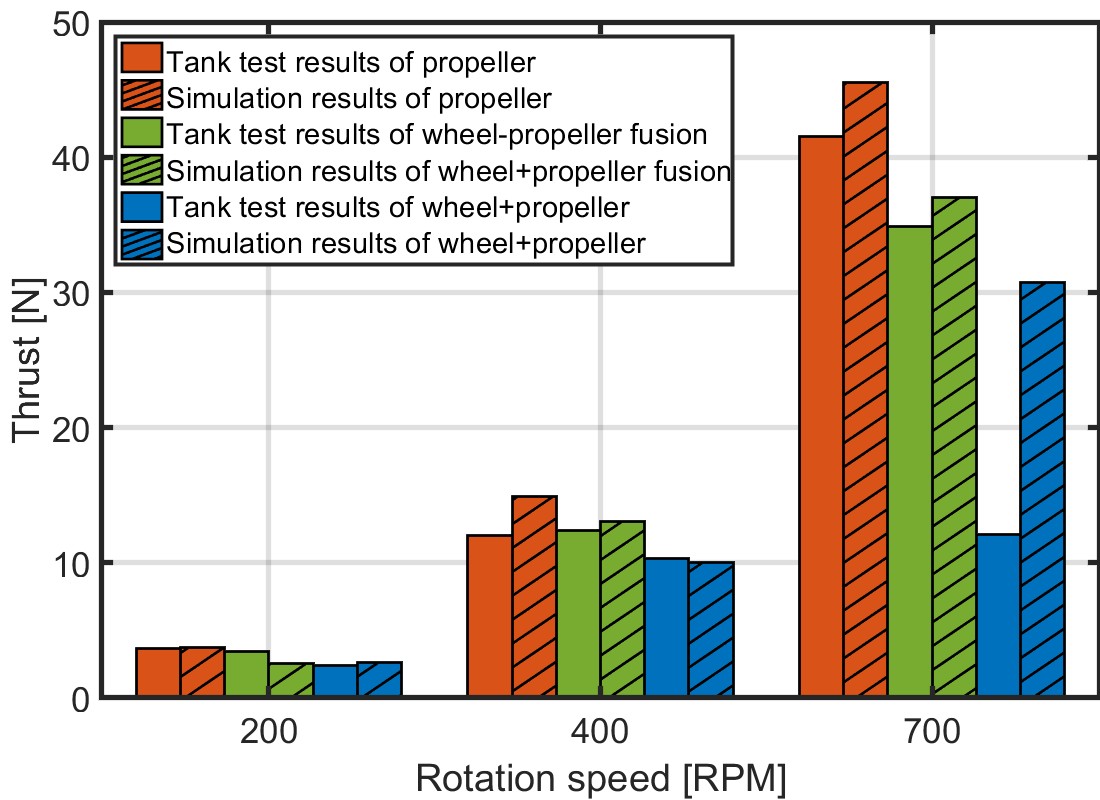}
  \caption{The Verification of the simulation results in the water tank.}
  \vspace{-5 mm} 
  \label{bartype}
\end{figure}

Simulated results in Fig. \ref{bartype} matched the experimental trends, confirming that the direct combination significantly increased drag. To mitigate this, we redesigned the wheel to contribute to thrust generation. Inspired by inclined deflectors on underwater thrusters, we implemented an inward-tilted rim that also prevents ground interference. The structure was fabricated by 3D printing and evaluated through simulations and experiments.

\begin{table}[t]
\caption{Comparison of Simulated and Experimental wheel-propeller fusion Thrust Results at Varying Rotation Speeds}
\centering
\resizebox{\columnwidth}{!}{
\begin{tabular}{cccc}
\hline
Rotation speed&Simulation results&Experimental results&Error\\
(RPM)&(N)&(N)&(\%)\\
\hline
200&2.53&2.605&2.9\\
300&7.24&6.4&11.6\\
400&13.07&11.1&15.07\\
500&20.33&17.72&12.83\\
600&28.40&24.1&15.64\\
700&37.02&34.86&5.83\\
\hline
\end{tabular}
}
 \vspace{-3 mm} 
 \label{simuexperror}
\end{table}

\begin{table}[t]
\caption{Tank test Performance Comparison of Different Propeller shapes in Thrust and Efficiency}
\centering
\resizebox{\columnwidth}{!}{
\begin{tabular}{ccccc}
\hline
Propeller&Max thrust&Max efficiency&Max speed&Max drag\\
type&(N)&(N/W)&(RPM)&torque(NM)\\
\hline
D180mm&49.6&0.449&734&- \\
W+P&12.1&0.209&441&0.603 \\
\textbf{fusion}&\textbf{36.5}&\textbf{0.399}&\textbf{710}&\textbf{0.06} \\
\hline
\end{tabular}
}
 \vspace{-5 mm} 
\end{table}

As shown in Fig. \ref{bartype}, the fusion design significantly reduced drag torque and approached the performance of the pure propeller. Although some added drag remained, causing a 26.4\% reduction in peak thrust, the maximum thrust increased by 201.6\% and efficiency nearly doubled compared to the original configuration. These results demonstrate the effectiveness of the proposed design.

\textbf{3) Selection of the Optimal Wheel-Propeller Param:}
\begin{figure*}[!t]
  \centering
  \includegraphics[width=\textwidth]{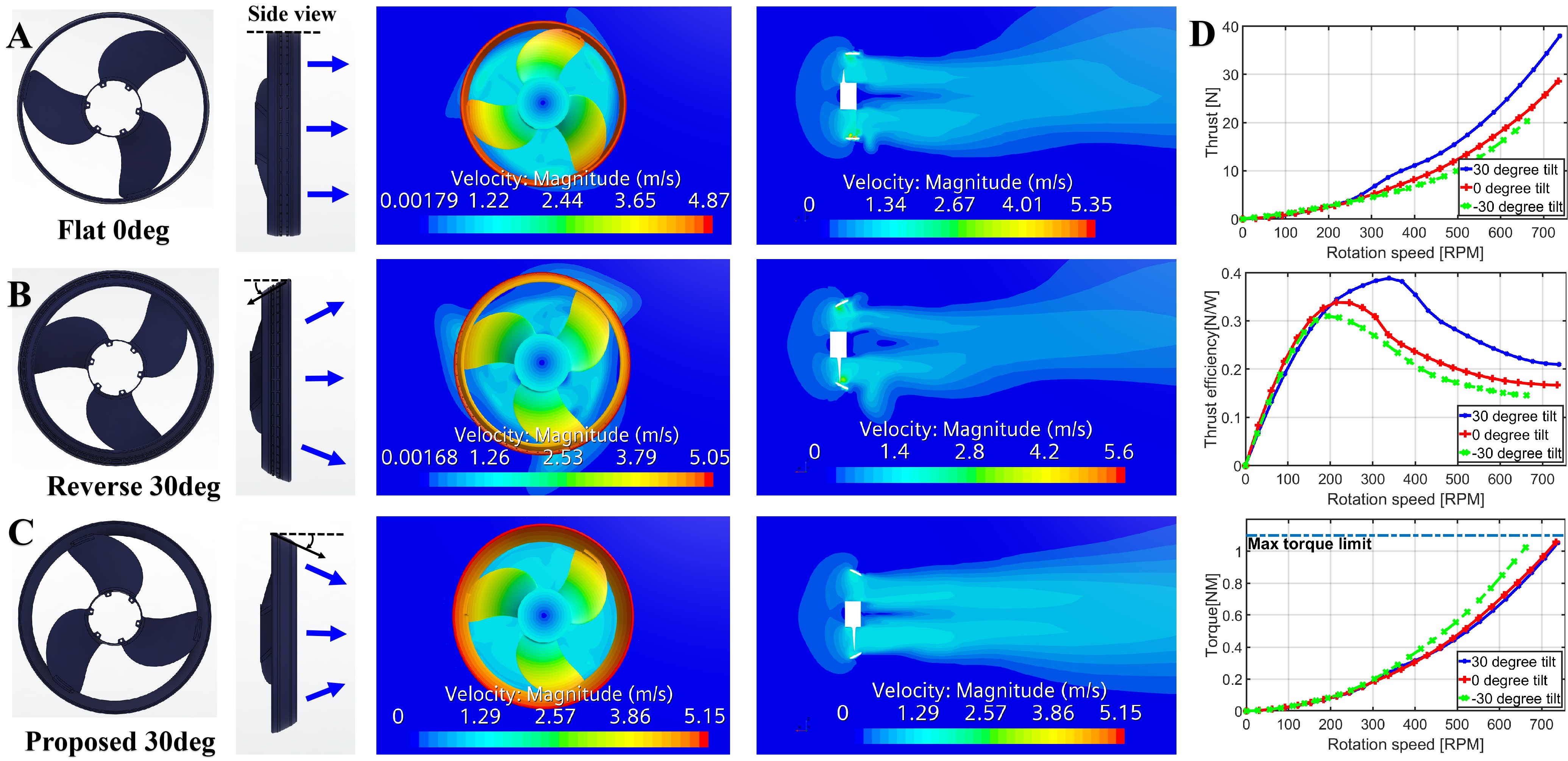}
    \caption{
    \textbf{A-C}.Comparison of hydrodynamic characteristics for different wheel–propeller tilt angle sets.
    \textbf{D}.Tank test results of three wheel–propeller tilt angle.
    }
    \vspace{-1 mm} 
    \label{simuangle}
\end{figure*}
\begin{figure*}[t]
  \centering
  \includegraphics[width=\textwidth]{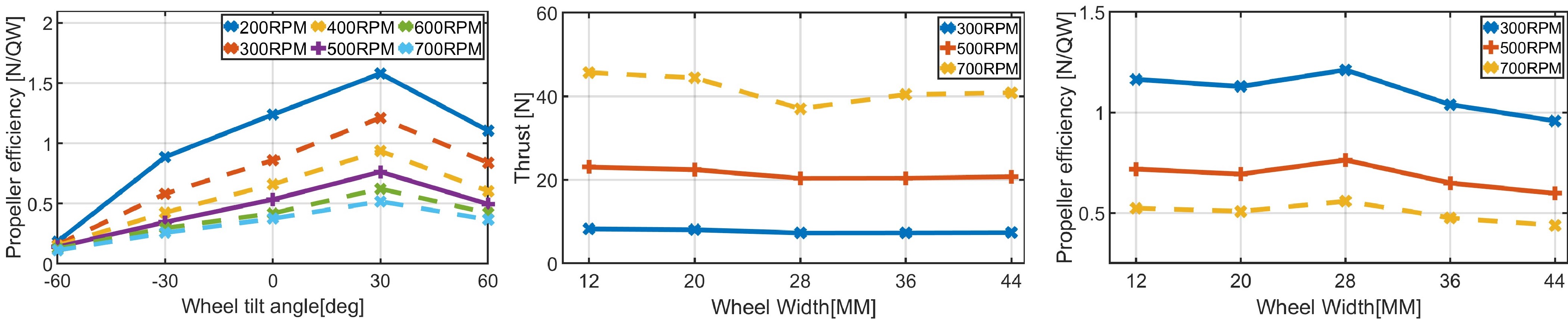}
  \caption{Simulation results of the thrust and efficiency in five different agnle at different rotate-speed and five different width at different speed.}
  \vspace{-3 mm} 
  \label{wanda}
\end{figure*}
\begin{figure}[t]
  \centering
  \includegraphics[width=\columnwidth]{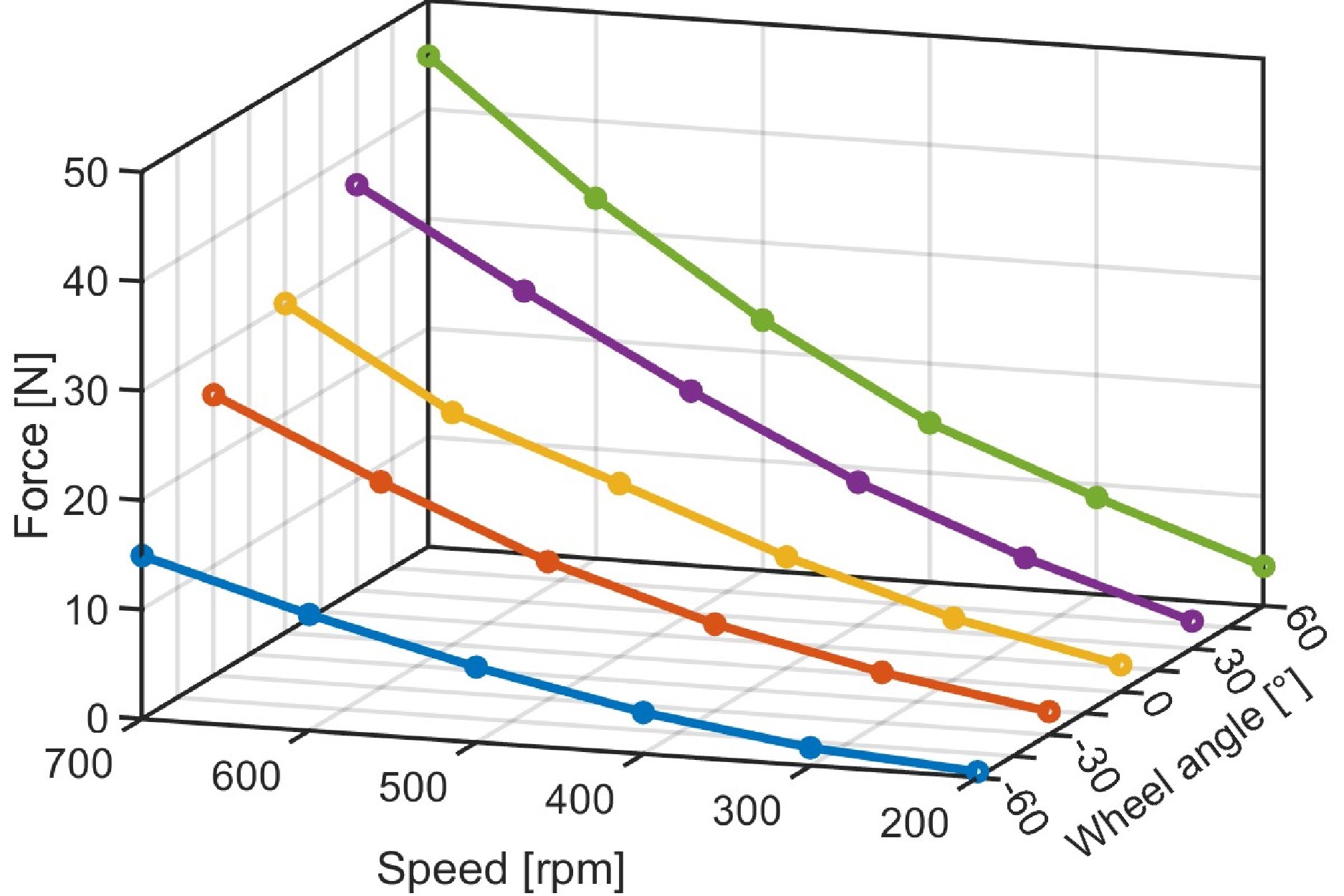}
  \caption{Simulation results of the thrust produced by the five different agnle at different rotate-speed.}
  \vspace{-3 mm} 
  \label{wange3d}
\end{figure}

Building on the success of the wheel–propeller fusion structure, we investigated the effect of wheel rim tilt angle on thrust and efficiency. Three configurations were tested: positive tilt, negative tilt, and horizontal, with simulations conducted across varying rotational speeds as shown in Fig. \ref{simuangle} and Fig. \ref{wange3d}.

As shown in Fig. \ref{wange3d} and Table \ref{anglecomtabe}, all configurations produced thrust above 300 r/min, but the 30 deg tilt consistently outperformed the others. At 600 r/min, it delivered 33\% higher thrust and 18\% higher efficiency than the horizontal setup. The -30 deg tilt suffered from excessive drag torque and performance drop-off. Flow field analysis (Fig. \ref{simuangle}.C) showed that the positive tilt generated a more concentrated wake, while the others produced more dispersed flows, leading to greater losses. Meanwhile as shown in Fig. \ref{wanda}, the efficiency first increases and then decreases with the tilt angle, reaching its peak at 30 degrees.

These results were validated through water tank experiments using 3D-printed prototypes, as shown in Fig. \ref{simuangle}.D and Table \ref{anglecomtabe}, confirming the superiority of the positive tilt design. Although increasing the tilt angle could further improve thrust, it would reduce ground clearance and compromise obstacle crossing ability. Therefore, the 30 deg positive tilt was chosen as the final configuration, balancing thrust enhancement with structural feasibility.

After determining the tilt angle, we further investigated the effect of wheel parameters on thrust and efficiency. Inspired by ducted propellers, we hypothesized that wheel width plays a key role. We tested five widths in 8 mm increments at three speed. 
As shown in Fig. \ref{wanda}, thrust remained stable at 300 and 500 rpm and showed a dip followed by an increase at 700 rpm, with the lowest thrust at 28 mm. In terms of efficiency, all speeds showed a rise then fall trend, peaking at 28 mm. Since thrust differences were small across widths except at 700 rpm, we prioritized efficiency and selected 28 mm. This sacrifices peak thrust for better overall performance and reduced weight. The performance is shown in Table \ref{anglecomtabe}.

\begin{table}[t]
\caption{Tank test Performance Comparison of Different wheel angles in Thrust and Efficiency}
\centering
\resizebox{\columnwidth}{!}{
\begin{tabular}{cccccc}
\hline
Propeller&Max&Test 600r&Simulation&Error&Efficiency\\
type&thrust(N)&thrust(N)&thrust(N)&(\%)&(N/W)\\
\hline
D180mm&49.6&24.98&29.39&15.01&0.449\\

\textbf{30deg}&\textbf{38.0}&\textbf{24.10}&\textbf{28.40}&\textbf{15.14}&\textbf{0.399} \\

0deg&28.6&18.12&21.21&14.56&0.338\\

-30deg&20.3&15.87&18.82&15.67&0.310\\
\hline
\end{tabular}
}
\label{anglecomtabe}
 \vspace{-5 mm} 
\end{table}

\textbf{4) Selection of the Optimal Wheel-Propeller Strips Number:}
The friction between the wheel–propeller and the ground can be enhanced by increasing the number of rubber strips, as described in (\cite{ma2022design}) has been well study. In our design, we applied rubber strips to the regions of the wheel rim that are likely to contact the ground. Additionally, tread patterns were incorporated into the 3D-printed wheel surface to improve traction and ensure effective performance in ground mode.

\begin{figure*}[!t]
  \centering
  \includegraphics[width=\textwidth]{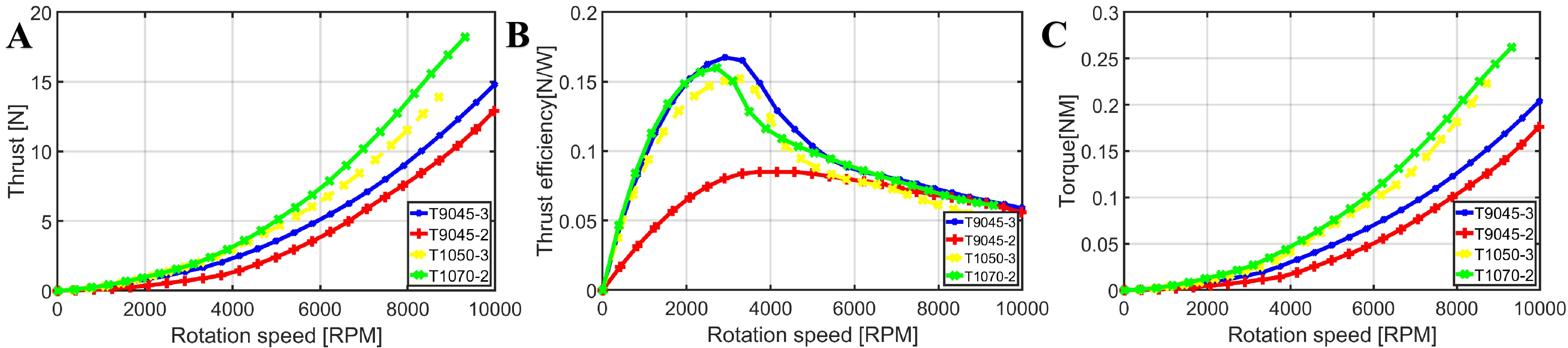}
  \caption{Performance evaluation of air-propellers in different diameter and blades number.
    \textbf{A}. The thrust-rpm curves.
    \textbf{B}. The efficiency-rpm curves. 
    \textbf{C}. The torque-rpm curves.}
  \label{airppall}
  \vspace{-2 mm}
\end{figure*}

\begin{table}[t]
\caption{Simulation Performance Comparison of Different width in Thrust and Efficiency}
\centering
\resizebox{\columnwidth}{!}{
\begin{tabular}{cccccc}
\hline
Width&Max&Max&300r&500r&700r\\
(mm)&thrust(N)&torque(NM)&Eff(N/W)&Eff(N/W)&Eff(N/W)\\
\hline
12&\textbf{45.6}&1.18&1.16&0.72&0.52\\
20&44.4&1.19&1.13&0.69&0.50 \\
\textbf{28}&37.0&\textbf{0.90}&\textbf{1.21}&\textbf{0.76}&\textbf{0.56}\\
36&40.4&1.16&1.042&0.65&0.48\\
44&40.8&1.27&0.96&0.60&0.44\\
\hline
\end{tabular}
}
\label{anglecomtabe}
 \vspace{-1 mm} 
\end{table}

\begin{table}[t]
\caption{Performance Comparison of Different Propeller size in Thrust and Efficiency}
\centering
\resizebox{\columnwidth}{!}{
\begin{tabular}{ccccc}
\hline
Propeller&Max thrust&Max efficiency&Valid efficiency(6000 r/min)\\
type&(N)&(N/W)&(N/W)\\
\hline
\textbf{T9045-3}&14.8&\textbf{0.1674}&0.0869\\
\textbf{T9045-2}&12.9&0.0849&0.0791\\
\textbf{T1050-3}&\textbf{18.2}&0.1599&\textbf{0.0882}\\
\textbf{T1070-2}&13.9&0.1519&0.0786\\
\hline
\end{tabular}
}
 \label{airpptable}
 \vspace{-2 mm} 
\end{table}

\subsection{2.4 Air propellers analysis and selection}
\textbf{1) Optimal propeller parameter selection:}
After finalizing the parameters of the integrated wheel–propeller structure, we proceeded to select a suitable aerial propeller. The chosen dual-output-shaft motor was originally designed for fixed-wing applications, and one-way bearings were added to the shaft housing to enable bidirectional operation. To identify the optimal propeller for the module, we tested four different propeller configurations varying in diameter and blade number. The evaluation focused on maximizing thrust and efficiency across the full throttle range (0\%–100\%).

As shown in Fig. \ref{airppall} and Table \ref{airpptable}, the maximum thrust increases with propeller size, and for the same diameter, three-blade propellers consistently outperform two-blade counterparts. In terms of efficiency, all configurations exhibit a typical trend: efficiency increases with rotation speed, reaches a peak around 3000 r/min, and then gradually declines. Most efficiency values cluster around 0.15 N/W. Beyond 6000 r/min, the 9-inch three-blade propeller achieves the highest efficiency, followed closely by the 10-inch three-blade variant, with a negligible difference of less than 0.01 N/W. However, the 9-inch propeller’s maximum thrust reaches only 81.7\% of the 10-inch model. Considering both thrust and efficiency, the 10-inch three-blade propeller offers the most favorable performance trade-off and was therefore selected for our system.

\begin{figure}[!t]
  \centering
  \includegraphics[width=\columnwidth]{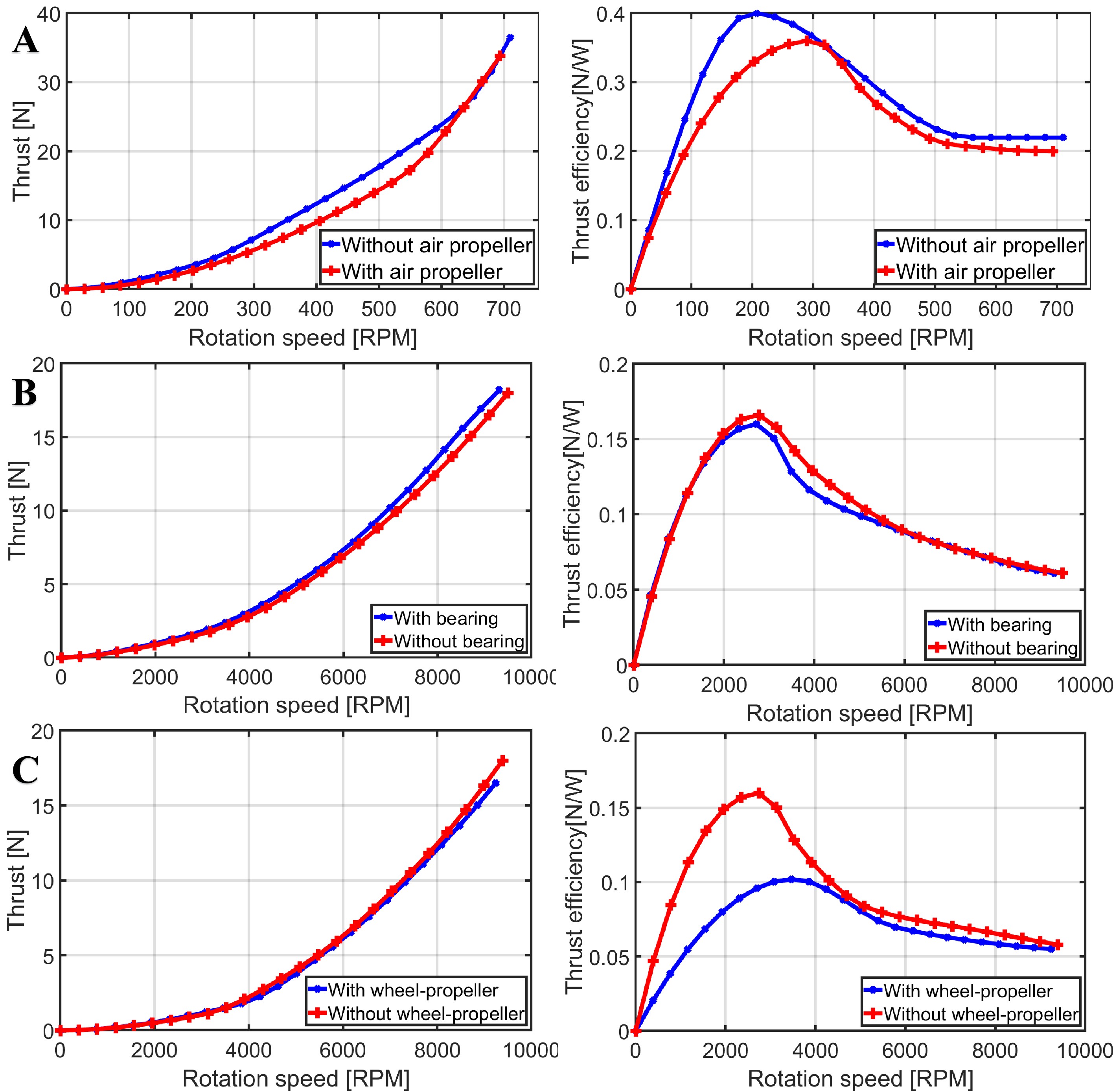}
  \caption{All-in-one propulsion unit performance evaluation experiments.
  \textbf{A}. Evaluation of aerial propeller interference on underwater mode.
  \textbf{B}. Evaluation of one-Way bearing impact on aerial mode.
  \textbf{C}. Evaluation of wheel–propeller interference with aerial inflow.}
  \label{distuall}
  \vspace{-4 mm}
\end{figure}

\textbf{2) Analysis of the interference of air propellers with wheel-propellers:}
After finalizing the propeller configurations, we evaluated the all-in-one propulsion unit to assess the performance losses caused by structural interactions. The detailed comparison is presented in Table \ref{testtable}.

Test 1: In underwater mode, the aerial propeller passively rotates and may induce hydrodynamic disturbance. As shown in Fig. \ref{distuall}.A, installing the aerial propeller led to a maximum thrust loss of 270 g (7.397\%) and an efficiency drop of 9.09\% -indicating limited interference.

Test 2: A one-way bearing was added to enable single-motor reuse in aerial mode. Fig. \ref{distuall}.B shows that this addition caused only 150 g (0.82\%) thrust loss, and efficiency degradation above 6000 r/min was negligible ($<$ 0.01 N/W), confirming the design’s effectiveness.

Test 3: In aerial mode, the wheel–propeller structure positioned above the aerial propeller may disrupt inflow. As shown in Fig. \ref{distuall}.C, this resulted in a thrust reduction of 150 g (8.3\%) and a 10.29\% efficiency drop at high speeds ($>$ 6000 r/min).

In conclusion, although the aerial mode incurs the highest efficiency loss, all observed degradations remain within acceptable limits. The results validate the integrated unit’s ability to support air, underwater, and ground propulsion modes with minimal trade-offs.
\begin{table*}[!t]
  \caption{Comparison of the Performance of Representative Drive Unit}
  \centering
  \resizebox{\textwidth}{!}
  {
  \begin{tabular}{c c c c c c c c c c c c}
  \hline
  Name&Weight&Aquatic Unit Efficiency&Aerial Unit Efficiency&Aerial Efficiency(N/W) &Aquatic Efficiency(N/W)&Max Aquatic Thrust/Weight Ratio&Air&Land&Water\\
  \hline
  \cite{liu2023tj}&122g&52.5\%&63.3\%&0.073&0.265&32N/122g = 0.262&\ding{51}& \ding{55} &\ding{51}\\
  \cite{tan2017efficient}&28g&46.0\%&50.0\%&0.031&0.172&9.5N/28g = 0.339&\ding{51}& \ding{55} &\ding{51}\\
  \cite{tan2020morphable}&19.3g&NA&12.3\%(simu)&NA&0.4(simu)&4.8N/19.3g = 0.249(simu)&\ding{51}& \ding{55} &\ding{51}\\  
  \cite{ma2022design}&$\ge$ 500g&NA&NA&NA&NA&55N/500g = 0.11&\ding{55}& \ding{51} &\ding{51}\\
  \cite{alzu2018loon}&130.4g&NA&NA&NA&0.18&12N/130.4g = 0.092&\ding{51}& \ding{55} &\ding{51}\\
  \cite{bi2022nezha}&83.4g&NA&NA&NA&0.27&15N/83.4g = 0.180&\ding{55}& \ding{55} &\ding{51}\\
  Proposed&\textbf{260g}&\textbf{48.8\%}&\textbf{72.73\%}&\textbf{0.083}&\textbf{0.36}&\textbf{34N/260g = 0.13} &\ding{51}& \ding{51} &\ding{51}\\
  \hline
  \end{tabular}
   }
\label{unitcompare}
\vspace{0 mm}
\end{table*}
\begin{figure*}[!t]
  \centering
  \includegraphics[width=\textwidth]{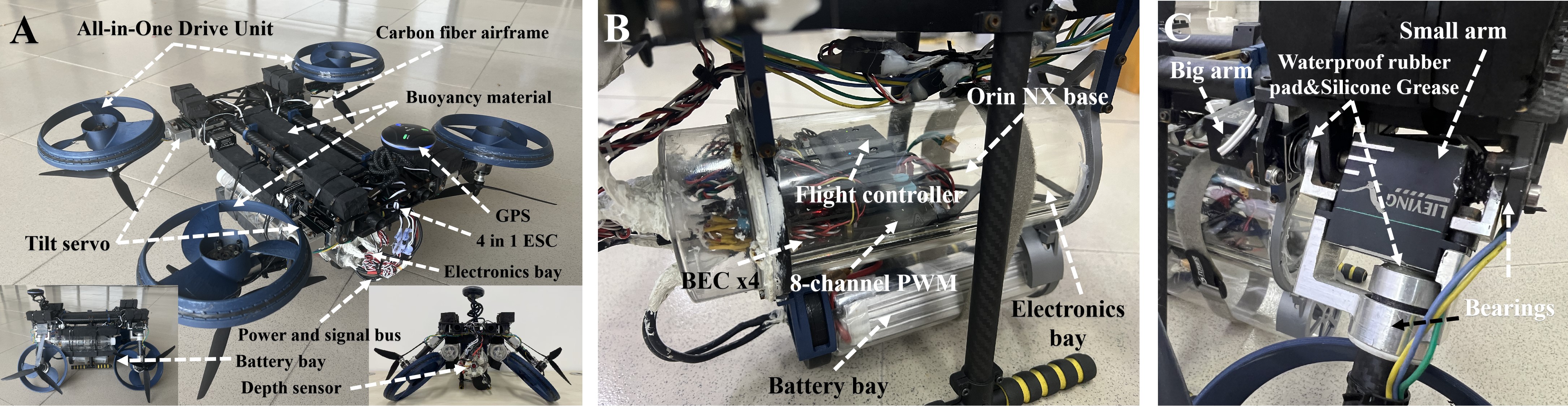}
  \caption{The prototype of Wukong-Omni and the labels of its main components.}
  \label{macpic}
    \vspace{-3 mm}
\end{figure*}
\begin{table}[t]
\caption{Comparison of interference between air propeller and wheel-propeller in Thrust and Efficiency}
\centering
\resizebox{\columnwidth}{!}{
\begin{tabular}{ccccc}
\hline
Exp&Max thrust&Proportion&Max efficiency&Proportion\\
type&loss(N)&(\%)&loss(N/W)&(\%)\\
\hline
\textbf{Test1}&2.7&7.397&0.02&9.09\\
\textbf{Test2}&2.0&1.098&0.0126&9.77\\
\textbf{Test3}&1.5&8.33&0.0074&11.2\\
\hline
\end{tabular}
}
 \label{testtable}
 \vspace{-4 mm} 
\end{table}
\subsection{2.5 Unit performance analysis}
To enhance the unit’s adaptability across multiple environments, we verified the effectiveness of the proposed design through both simulations and tank tests. We systematically investigated the effects of fusion structure and wheel tilt angle and width on performance. This configuration achieved a 201.6\% increase in maximum thrust and a 90.9\% improvement in peak efficiency compared to the simple combination of wheel and propeller.

Furthermore, ablation experiments were conducted to evaluate performance degradation when all subsystems operate simultaneously. Results showed that the maximum efficiency loss ranged from 1.098\% to 11.2\%. The proposed unit represents the first propulsion unit capable of operating seamlessly across air, water, and land. Compared with existing dual-domain (air–ground or air–water) system, although our design incurs a moderate weight penalty due to the integration of both aerial and wheel-propellers, it outperforms all existing systems in both aerial and underwater thrust-to-power efficiency. The only slight compromise lies in the underwater gear reduction stage, where efficiency is marginally lower than the best-reported value. These results underscore the exceptional performance of our design. The detailed comparison is summarized in Table \ref{unitcompare}.

\section{3 PROTOTYPE DESIGN}
\subsection{3.1 Mechanism Design}
To enable high maneuverability across diverse environments, Wukong-Omni adopts a quadrotor-based configuration with two additional rotational degrees of freedom at each arm root: the large arm folds about the roll axis, and the small arm rotates the propulsion module about its own axis. To ensure lightweight yet robust actuation, the joint connecting the upper and lower arms is fabricated from CNC-machined aluminum alloy and integrates dual deep-groove ball bearings, enabling precise, eccentricity-free rotation of the carbon tube. A custom connector couples the tube to the servo output for torque transmission, while retaining rings prevent axial displacement.
\textbf{\begin{table}[t]
  \caption{Prototype Component Selection}
  \centering
  \resizebox{0.8\columnwidth}{!}{
  \begin{tabular}{c c}
    \hline
  Component&Type \\
  \hline
  Motor&T-MOTOR AT2321-KV950\\
  Aerial propeller&GEMFAN1050X3\\
  Aerial ESCs&velox-V45A-V2\\
  Battery&Geshi LiPo 4S 5300mAh\\
  Flight Controller&Pixhawk6C\\
  Onboard Computer&Orin Nx-16Gb\\
  Camera Module&Oak-4P-New\\
  GPS& CUAV Neo3X\\
  Telemetry Radio&CUAV PW-LINK/Holybro 433MHz\\
  Transmitter&TBS CROSSFIRE-NANO\\
  Receiver&TBS NANO-RX\\
  BEC&Mayatech 5V3A\\
  Servo&LIEYING1025-60KG\\
 Depth Sensor&MS5837\\
 \hline
 \end{tabular}
 }
 \label{electable}
 \vspace{-3mm} 
\end{table}}

A clamping bracket mounts the electronics and battery compartments beneath the arms, lowering the center of gravity and increasing displacement during water emergence. Wukong-Omni is positively buoyant, with buoyancy blocks placed above the electronics to vertically align the centers of gravity and buoyancy, which keeping the latter higher for passive stability and capsize resistance. These blocks also assist in generating lift during transitions from water to air. The structural layout is shown in Fig.~\ref{macpic}.

Wukong-Omni weighs 4.4 kg. Despite the added mass from servos, the design enables robust and flexible operation across multiple domains. By optimizing buoyancy distribution and mechanical layout, Wukong-Omni achieves smooth inter-domain transitions and stable performance in air, on land, and underwater.
\begin{figure*}[!t]
  \centering
  \includegraphics[width=\textwidth]{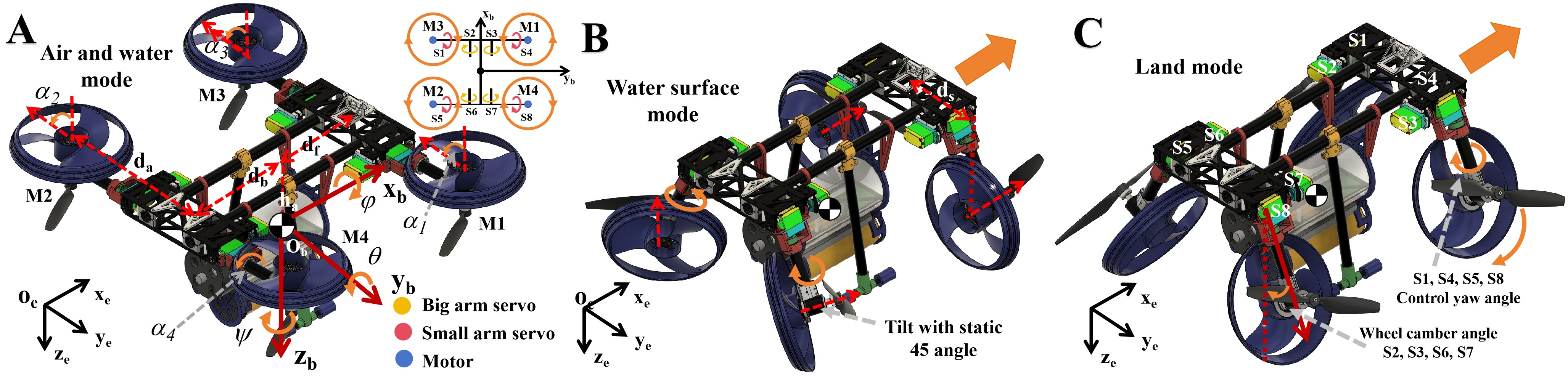}
  \caption{Model of Wukong-Omni with the labels on the coordinate systems and thrusters arrangement in air, land, water surface mode.}
  \label{modalpic}
    \vspace{-3 mm}
\end{figure*}
\begin{figure*}[!t]
  \centering
  \includegraphics[width=\textwidth]{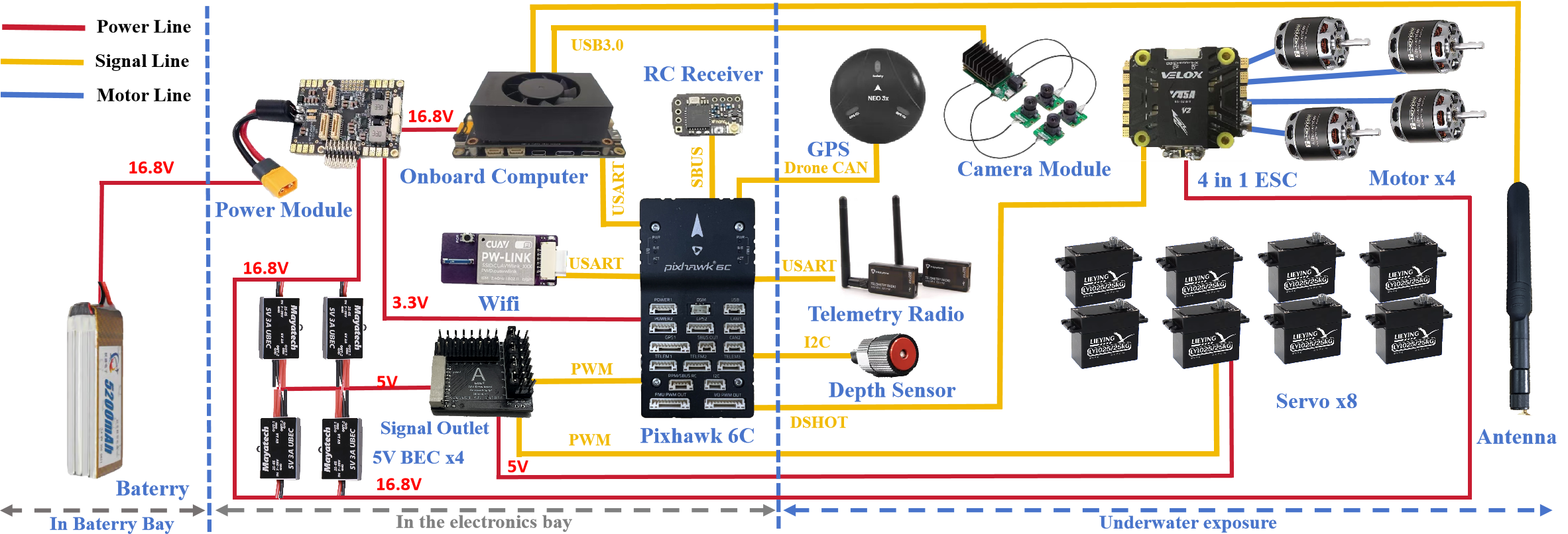}
  \caption{Electronic and hardware framework of Wukong-Omni.}
  \label{ELECPIC}
  \vspace{-4  mm}
\end{figure*}
\subsection{3.2 Avionics Component Selection}
As shown in Fig. \ref{ELECPIC}, Wukong-Omni's electronic system is built around a Pixhawk 6C flight controller running customized firmware. An onboard Orin NX handles perception and planning, connected to the flight controller via serial and linked to an omnidirectional fisheye camera for future tasks such as visual localization and navigation. But at this stage, they are omitted, since only structural and low-level control are evaluated.

To address underwater signal attenuation, an 868MHz TBS CROSSFIRE receiver provides strong penetration and up to 2W transmission power. Telemetry is supported by Wi-Fi and a 433MHz module, mainly used during gain tuning.

A water pressure sensor enables closed-loop depth control and assists in switching between modes. Four independent voltage regulators provide stable power to servos. As shown in Fig. \ref{macpic}, the battery is installed in a quick-release waterproof compartment for easy replacement, while other electronics are enclosed in the main bay. Waterproof silicone and grease are applied at key joints and rotating parts to ensure reliable operation in harsh environments. Details are listed in Table \ref{electable}.
\section{4 MULTI-MODAL DYNAMICS OF THE Wukong-Omni}
As illustrated in Fig. \ref{modalpic}.A, we define a right-hand inertial frame $\boldsymbol{E} = (O_e, x_e, y_e, z_e)$, with origin $O_e$ fixed on the water surface and the $z_e$-axis pointing vertically downward toward the Earth. The body-fixed frame of the vehicle is denoted by $\boldsymbol{B} = (O_b, x_b, y_b, z_b)$, where $O_b$ is located at the vehicle’s center of mass ($C_G$ in Fig. \ref{modalpic}.A).

The position of the vehicle in the inertial frame is represented by the vector $\boldsymbol{P} = [x \ y \ z]^T$, and its orientation is described by the Euler angle vector $\boldsymbol{\Theta} = [\phi \ \theta \ \psi]^T$. The linear velocity and angular velocity expressed in the body frame $\boldsymbol{B}$ are denoted as $\boldsymbol{\upsilon} = [u \ v \ w]^T$ and $\boldsymbol{\Omega} = [p \ q \ r]^T$, respectively. The rotation matrix $\boldsymbol{R}_B^E \in SO(3)$ transforms vectors from the body frame $\boldsymbol{B}$ to the inertial frame $\boldsymbol{E}$ is given by
 \begin{equation}
  \boldsymbol{R}_B^E = \begin{bmatrix} 
    c\theta c\psi & s\phi s\theta c\psi-c\phi s\psi& c\phi s\theta c\psi+s\phi s\psi
    \\ c\theta s\psi &s\phi s\theta s\psi + c\phi c\psi & c\phi s\theta s\psi - s\phi c\psi
    \\ -s\theta & s\phi c\theta & c\phi c\theta
    \end{bmatrix} .
  \end{equation}
The transformation matrix of angular rate $\boldsymbol{W}$ is given by
  \begin{equation}
    \boldsymbol{W} = \begin{bmatrix} 
      1 & s\phi t\theta & c\phi t\theta
      \\ 0 & c\phi & -s\phi
      \\ 0 & s\phi/c\theta & c\phi/c\theta
      \end{bmatrix} ,
    \end{equation}
where $c(\cdot)$, $s(\cdot)$and $t(\cdot)$ represent the abbreviation of cos$(\cdot)$ sin$(\cdot)$ and tan$(\cdot)$. Then the kinematic equation of the robot can be expressed as
\begin{equation}
  \begin{aligned}
    \dot{P} = \boldsymbol{R}_B^E \upsilon, \quad \dot{\Theta} = \boldsymbol{W} \Omega.
  \end{aligned}
\end{equation} 
To reduce computational complexity and facilitate controller implementation, coupling and Coriolis effects are omitted given the system’s lightweight nature and limited processing capacity. The resulting multi-domain dynamics in the body-fixed frame $\boldsymbol{B}$ is given by
\begin{align}
    (\boldsymbol{J} + \boldsymbol{J}_{a}) \boldsymbol{\dot{\Omega}} &= \boldsymbol{M}_{\mathrm{all}} -\boldsymbol{\Omega} \times (\boldsymbol{J} + \boldsymbol{J}_{a}) \boldsymbol{\Omega} - \boldsymbol{M}_{\mathrm{drag}} ,
  \\
    (m+\widetilde{m}_{a})\boldsymbol{\dot{\upsilon}} &= -(m+\widetilde{m}_{a}) \boldsymbol{\upsilon} \times \boldsymbol{\Omega} +\boldsymbol{F}_{\mathrm{all}} \nonumber \\
    &+ (m+\widetilde{m}_{a})\boldsymbol{g} - \boldsymbol{F}_{\mathrm{drag}} - \boldsymbol{F}_{b} + \boldsymbol{F}_{n},
\end{align}
where $\boldsymbol{F}_{all}$ and $\boldsymbol{M}_{all}$ represent the control force and torque across multiple domains, respectively. When robot runs in the land, we assume that the land is flat. Therefore we assume the support force $\boldsymbol{F}_n$ is completely the same as the robot weight.
The variable $m$ denotes the robot’s mass, and $\boldsymbol{J} = \mathrm{diag}([I_{xx}, I_{yy}, I_{zz}])$ is its inertia matrix. The variable $k$ is a mode flag, with $k = 0$, 1, 2, and 3 corresponding to aerial, land, underwater, and surface operation modes, respectively. The hydrodynamic model follows the formulation presented in (\cite{liu2024wukong}), and the added inertia $\boldsymbol{J}_a$ and added mass $\widetilde{m}_a$ are expressed as
\begin{align}
\widetilde{m}_{a} &= 
\begin{cases}
0, &{(k=0,1)}\\[3pt]
m_{a},&{(k=2,3)}
\end{cases}
\\[4pt]
\boldsymbol{J}_a &= 
\begin{cases}
\boldsymbol{0}_{3 \times 3}, & {(k=0,1)} \\[3pt]
\mathrm{diag}
\bigl( \bigl[
J_{ax}, J_{ay}, J_{az}
\bigr] \bigr), & {(k=2,3)}
\end{cases}
\end{align}
where the added mass and added inertia parameters can be calculated. To facilitate this, the robot is simplified as a cuboid, allowing the parameters to be expressed as
\begin{align}
{m}_a &= k_{am}\rho_w L_a W_a H_a,\\[3pt]
    \begin{bmatrix}
    J_{ax}\\
    J_{ay}\\
    J_{az}
    \end{bmatrix}&= \frac{1}{12} m_a
\begin{bmatrix}
W_a^2 + H_a^2 \\
L_a^2 + H_a^2 \\
L_a^2 + W_a^2
\end{bmatrix},
\end{align}
where $L_a$, $W_a$, and $H_a$ represent the robot’s length, width, and height, respectively. The drag force $\boldsymbol{F}_{\text{drag}}$ and drag torque $\boldsymbol{M}_{\text{drag}}$ can be expressed as
\begin{align}
\boldsymbol{M}_{drag} &= 
\begin{cases}
\bigl[
0, 0, 0
\bigr]^T, &{(k=0,1)}\\[4pt]
\boldsymbol{K}_{d} \left\| \boldsymbol{\Omega} \right\| \boldsymbol{\Omega},&{(k=2,3)}
\end{cases}
\\[4pt]
\boldsymbol{F}_{drag} &= 
\begin{cases}
\bigl[
0, 0, 0
\bigr]^T, &{(k=0,1)}\\[4pt]
0.5 \rho_{w} C_{{d}} \left\| \boldsymbol{\upsilon} \right\| \boldsymbol{\upsilon},&{(k=2,3)}
\end{cases}
\hspace{7mm}
\end{align}
where $\rho_{w}$ is the water density, $C_{d}$ is the drag coefficient, $\boldsymbol{K}_d$ is the drag moment coefficient, and the buoyant force $\boldsymbol{F_b}$ is expressed as
\begin{equation}
\begin{alignedat}{2}
\boldsymbol{F}_{b} &= 
\begin{cases}
\bigl[
0, 0, 0
\bigr]^T, &{(k=0,1)}\\[4pt]
\rho_{w} V \boldsymbol{g},&{(k=2,3)}
\end{cases}
\end{alignedat},
\hspace{15mm}
\end{equation}
where $V$ is the volume of the robot, and $\boldsymbol{g} = [0\ 0\ g]^T$ is the gravitational acceleration vector. and the control force and control torque generated by the all-in-one propulsion unit across multiple domains are expressed as
\begin{equation}
\boldsymbol{F}_{all}=\begin{cases}
\boldsymbol{F}_{a} &{(k=0)}\\[4pt]
\boldsymbol{F}_{l}&{(k=1)}\\[4pt]
\boldsymbol{F}_{w}&{(k=2)}\\[4pt]
\boldsymbol{F}_{s}&{(k=3)}
\end{cases},
\boldsymbol{M}_{all}=\begin{cases}
\boldsymbol{M}_{a} &{(k=0)}\\[4pt]
\boldsymbol{M}_{l}&{(k=1)}\\[4pt]
\boldsymbol{M}_{w}&{(k=2)}\\[4pt]
\boldsymbol{M}_{s}&{(k=3)}
\end{cases},
\end{equation}
the control forces and torques in different domains will be introduced in the following parts.
When the unit operates in different modes, the generated thrust and torque can be expressed as
\begin{align}
    T_{ji} &= K_{Tja} \omega_{ji}^2 \! + \! K_{Tjb} \omega_{ji} \! + \! K_{Tjc}, \quad M_{ji} = K_{Qj} T_{ji}, \nonumber \\
    i &= 1, 2, 3, 4, \quad j = A(air),M(water),L(land),
\end{align} 
The positive coefficients $K_{Tja-c}$ and $K_{Qj}$ represent the thrust and torque factors, respectively, determined by the thruster's mechanical properties. Bidirectional thrust can be generated underwater by reversing the rotation direction.

\textbf{1) The underwater mode control allocation model:}
The thruster layout is shown in Fig. \ref{modalpic}.A. Here, $d_a$ is the horizontal distance from the center of mass to the aerial thrusters, while $d_f$ and $d_b$ are the vertical distances to the front and rear thruster pairs. $\alpha_1$ to $\alpha_4$ denote the tilt angles of the aerial thrusters. Underwater, a conventional quadrotor control allocation model is used, enabling high maneuverability. The corresponding underwater model is given by:
\begin{align}
   \boldsymbol{F}_{w} \hspace{-1mm} = &  \hspace{-1mm} 
    \begin{bmatrix}
    \boldsymbol{0}_{2 \times 1} \\ 
    T_{M1} + T_{M2} + T_{M3} + T_{M4}
    \end{bmatrix}, \\
    \boldsymbol{M}_{w}\hspace{-1mm} = &\hspace{-1mm}  
    \begin{bmatrix}
     d_a & -d_a & -d_a & d_a\\
   -d_f & d_b & -d_f & d_b \\
    -s\alpha_1 d_a & \hspace{-1mm}s\alpha_2 d_a &\hspace{-1mm} s\alpha_3 d_a & \hspace{-1mm}-s\alpha_4 d_a\\
    \end{bmatrix}
    \hspace{-1mm}\hspace{-1mm}
        \begin{bmatrix}
      T_{M1}\\T_{M2}\\T_{M3}\\T_{M4}
      \end{bmatrix}
    \label{equ:control_allocatewater},
    \end{align}
    
\textbf{2) The air mode control allocation model:}
In aerial mode, the vehicle adopts a tilting rotor configuration. Yaw torque is generated by tilting the rotors, which significantly enhances control authority around the yaw axis. Meanwhile, roll and pitch torques are realized through differential thrust between opposing rotors. As illustrated in Fig. \ref{modalpic}.A, the coordinate system is defined with respect to the center of mass. Based on this configuration, the dynamic models are formulated as follows
\begin{align}
    \boldsymbol{F}_{a}\hspace{-1mm} &= \hspace{-1mm} 
    \begin{bmatrix}
        \boldsymbol{0}_{2 \times 1} \\
        -T_{A1} - T_{A2} - T_{A3} - T_{A4}
    \end{bmatrix},
    \\\boldsymbol{M}_{a} \hspace{-1mm}&=  \hspace{-1mm}
    \begin{bmatrix}
       -d_a &  d_a &  d_a & - d_a \\
       d_f & -d_b & d_f & -d_b \\
     s\alpha_1 d_a & \hspace{-1mm}-s\alpha_2 d_a & \hspace{-1mm}-s\alpha_3 d_a & \hspace{-1mm}s\alpha_4 d_a
      \end{bmatrix}
    \begin{bmatrix}
      T_{A1}\\T_{A2}\\T_{A3}\\T_{A4}
      \end{bmatrix},
\end{align}

\textbf{3) The water surface mode control allocation model:}In the water surface mode, a boat-like propulsion configuration is adopted. Due to mechanical constraints, the rear rotors are limited to a maximum tilt angle of 45 deg, which is utilized during high-speed forward motion. Yaw control is achieved via differential thrust, while pitch and roll are passively stabilized by buoyancy and hull design. The coordinate system is defined in Fig.~\ref{modalpic}.B, and the corresponding control models are expressed as
\begin{align}
    \boldsymbol{F}_{s} \hspace{-1mm}&=\hspace{-1mm}  
    \begin{bmatrix}
    T_{M1} + T_{M3} + \frac{\sqrt{2}}{2}T_{M4} + \frac{\sqrt{2}}{2}T_{M2}\\
      \frac{\sqrt{2}}{2}T_{M4} -  T_{M2}\frac{\sqrt{2}}{2} \\
        0
    \end{bmatrix},
    \\\boldsymbol{M}_{s} \hspace{-1mm}&=\hspace{-1mm}  
    \begin{bmatrix}
        \boldsymbol{0}_{2 \times 1} \\
        (T_{M3} \hspace{-1mm}-\hspace{-1mm} T_{M1})d_{s}\hspace{-1mm}+\hspace{-1mm}\frac{\sqrt{2}}{2}(T_{M2} \hspace{-1mm}- \hspace{-1mm}T_{M4})(d_{s}\hspace{-1mm}+\hspace{-1mm}d_s) 
    \end{bmatrix}.
\end{align}
\begin{figure*}[!t]
  \centering
  \includegraphics[width=\textwidth]{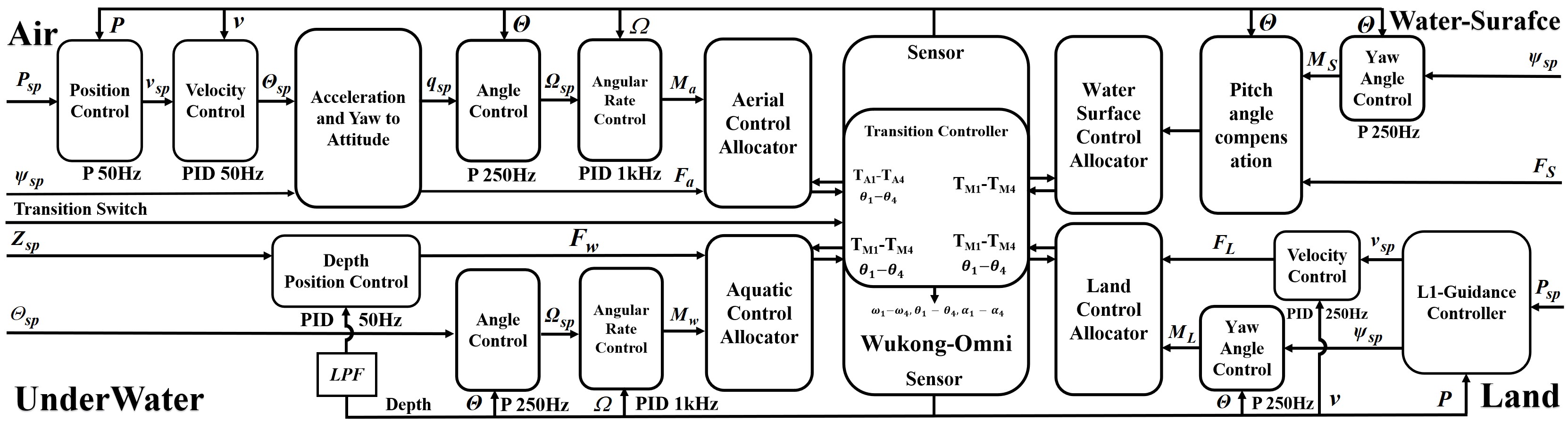}
  \caption{Wukong-Omni’s integrated control framework structure.}
  \label{controlstar}
   \vspace{-5mm} 
\end{figure*}

\textbf{4) The land mode control allocation model:}
In land mode, the system simplifies from a 6-DOF to a 2-DOF model. For tractability, effects like rolling resistance and gyroscopic moments are neglected. Assuming sufficient ground traction, the vehicle is modeled as a rigid body on a flat surface. A step function $H(x)$ captures directional switching: due to a one-way bearing, only the front wheels and steering servo are active during forward motion, while the rear wheels engage in reverse. The coordinate system is shown in Fig. \ref{modalpic}.C, and the resulting control allocation model is
\begin{align}
    \boldsymbol{F}_{l} \! &= \!
    \begin{bmatrix}
      T_{L1}H(u)\hspace{-1mm}-\hspace{-1mm}T_{L2} H(-u)\hspace{-1mm}+\hspace{-1mm}T_{L3} H(u)\hspace{-1mm}-\hspace{-1mm}T_{L4} H(-u)\\
      \boldsymbol{0}_{2 \times 1}
      \end{bmatrix}\hspace{-1mm},
      \nonumber \\
      \boldsymbol{M}_{l} \! &= \!
    \begin{bmatrix}
       0 & 0 &  0 & 0 \\
       0 & 0 & 0 & 0 \\
     H(u) & \hspace{-1mm}-H(-u) & \hspace{-1mm}-H(u) & \hspace{-1mm}H(-u)
      \end{bmatrix}
      \!
      \begin{bmatrix}
      \alpha_1\\\alpha_2\\\alpha_3\\\alpha_4
      \end{bmatrix},
\end{align}
\section{5 CONTROL AND TRANSITION STRATEGY}
\subsection{5.1 The multi-mode control law design }
The whole hybrid control structure as Fig. \ref{controlstar} shows. On land, the robot operates within a 2D plane, with yaw and forward velocity control using GPS feedback. Wukong-Omni employs the L1 guidance law for autonomous navigation. When a waypoint is assigned, the robot approaches the target along a path that approximates the smallest feasible inscribed circle. The desired lateral acceleration is calculated as follows
\begin{align}
a_d = &\frac{4 \cdot \zeta^2 \cdot u^2 \cdot \sin(\epsilon_1  + \epsilon_2)}{d_{L1}}, \\
&d_{L1}=\frac{1}{\pi} \zeta T_{L1}^2u,
\end{align}
where $\zeta$ is the damping ratio, $T_{L1}$ is the L1 control period, $u$ is the velocity, $d_{L1}$ is the look-ahead distance, $\epsilon_1$ is the angle from the robot to $d_{L1}$ point vector to waypoint vector, and $\epsilon_2$ is the velocity angle relative to the waypoint vector. The required steering angle is then computed based on the ackermann kinematics and centripetal acceleration formula follows
\begin{equation}
\theta = \frac{\pi}{2} - \arctan\left(\frac{u^2 }{a_dL_c}\right),
\end{equation}
where $L_c$ is the wheelbase of the robot. This guidance strategy is also used in position control for fixed-wing UAVs (\cite{park2004new}).

In aerial mode, Wukong-Omni utilizes a cascaded PID control strategy. Underwater, where position feedback is unavailable, the user manually inputs the target attitude. A cascaded PID controller also manages the attitude loop, with the angular velocity controller running at 1kHz to ensure rapid response. Depth, measured via a pressure sensor, is control by PID controller operating at 50Hz. To reduce noise, the depth signal is passed through a 13Hz Butterworth low-pass filter, enabling Wukong-Omni to maintain stable underwater depth.

On the water surface, control remains two-dimensional. The user inputs only the desired yaw and forward thrust, which similar to marine vessel operation. The all model parameters are set as shown in TABLE \ref{modelparam}.

\subsection{5.2 Cross Domain Transition Strategy}
In domain transition strategy, a transition controller is employed to manage outputs from four operational modes, as illustrated in Fig. \ref{controlstar}. This controller receives control commands from the control allocators, along with selection signals from the remote switch. When a specific mode is activated, its corresponding control outputs are executed, while the other mode output are temporarily disabled to avoid conflicts.

To ensure seamless transitions, the switching controller also maintains a finite state machine that handles mode-specific initialization tasks, such as presetting joint angles. 
\subsubsection{\textbf{1) Analysis of Multi-mode Constraints on Transition:}}
Because the all-in-one propulsion unit must operate across multiple domains, the environmental constraints differ significantly. In aerial mode, the propulsion unit points upward and experiences no interference. In contrast, in land mode, the folded arm configuration may cause the aerial propeller to strike the ground, hindering proper operation.

\begin{figure}[t]
  \centering
  \includegraphics[width=0.95\columnwidth]{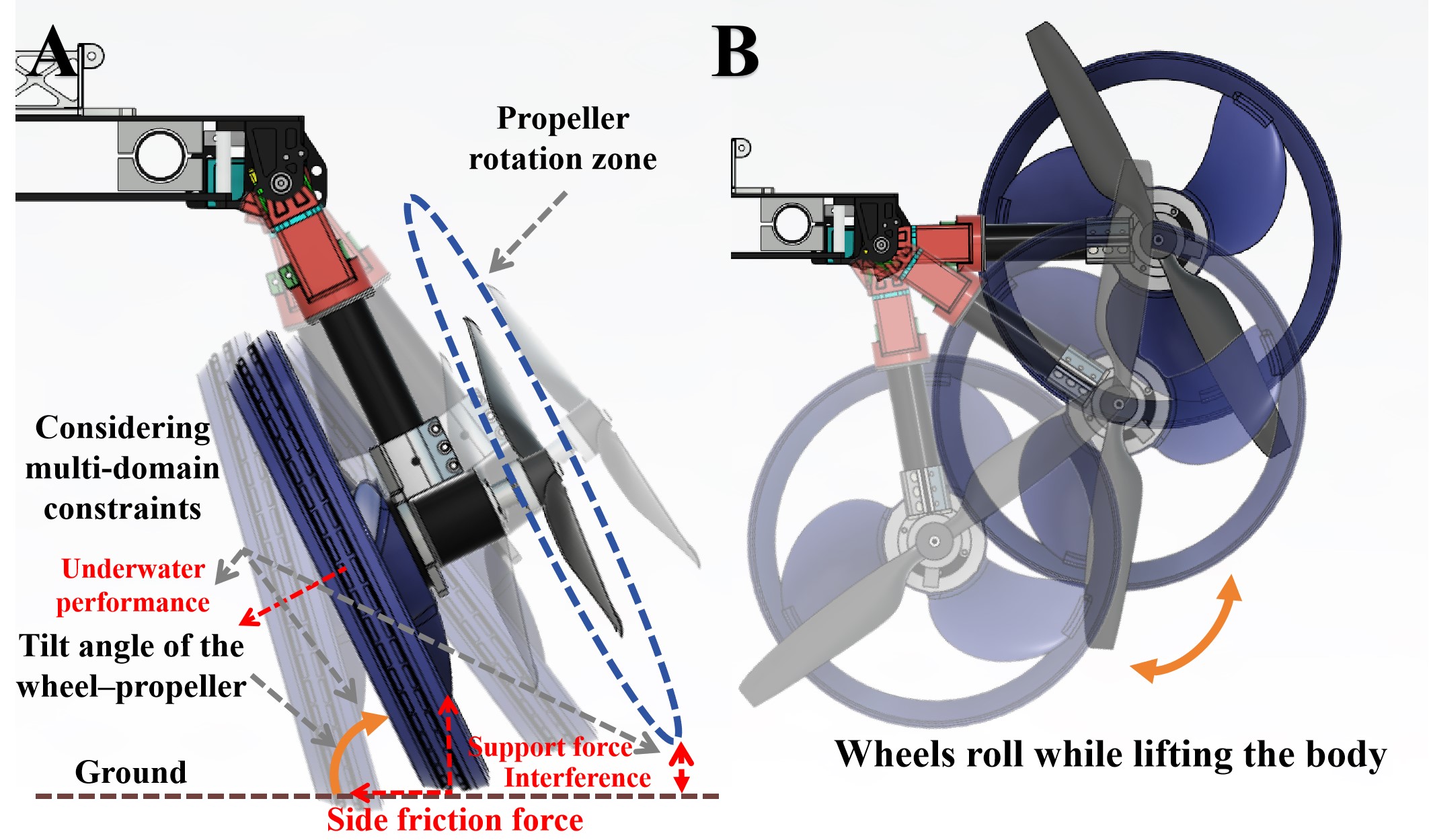}
  \caption{ 
  \textbf{A}. Multi-constraint analysis of wheel–propeller configuration and tilt angle. \textbf{B}. Illustration of main-arm rotation during the air-to-ground transition.}
  \label{wpstrapic}
   \vspace{-5 mm} 
\end{figure}

The main challenge is the ground mode, as illustrated in Fig. \ref{wpstrapic}.A.

First, in a conventional Ackermann chassis, the wheels stand nearly vertical to the ground, causing the aerial propeller to collide with the surface during motion. Reducing the propeller diameter could mitigate this, but earlier analysis shows that doing so greatly reduces thrust and efficiency.

Second, introducing a wheel–propeller tilt angle can enhance underwater thrust and help avoid propeller–ground interference. In our design, the wheel tilts with the propeller, so tire contact remains consistent. However, a larger tilt redirects part of the supporting normal force laterally and reduces the effective normal force for traction, increasing the required lateral friction and altering ground force balance. These changes can degrade steering stability, raise rolling resistance, and limit land mode mobility.

Third, without tilt, neither underwater thrust nor efficiency can reach their optimal values, and the aerial propeller would still interfere with the ground.

Considering all three factors, a particular configuration is required which can balances underwater propulsion performance, ground-mode traction, and propeller–ground clearance. Using the previously optimized 30 degree wheel–propeller tilt as the baseline, we further adjust the main-arm angle to determine the minimum tilt required to avoid ground interference without compromising ground-mode friction and terrain passability. Experiments verify that this configuration enables reliable and consistent operation across all three modes.

\begin{figure}[t]
  \centering
  \includegraphics[width=0.95\columnwidth]{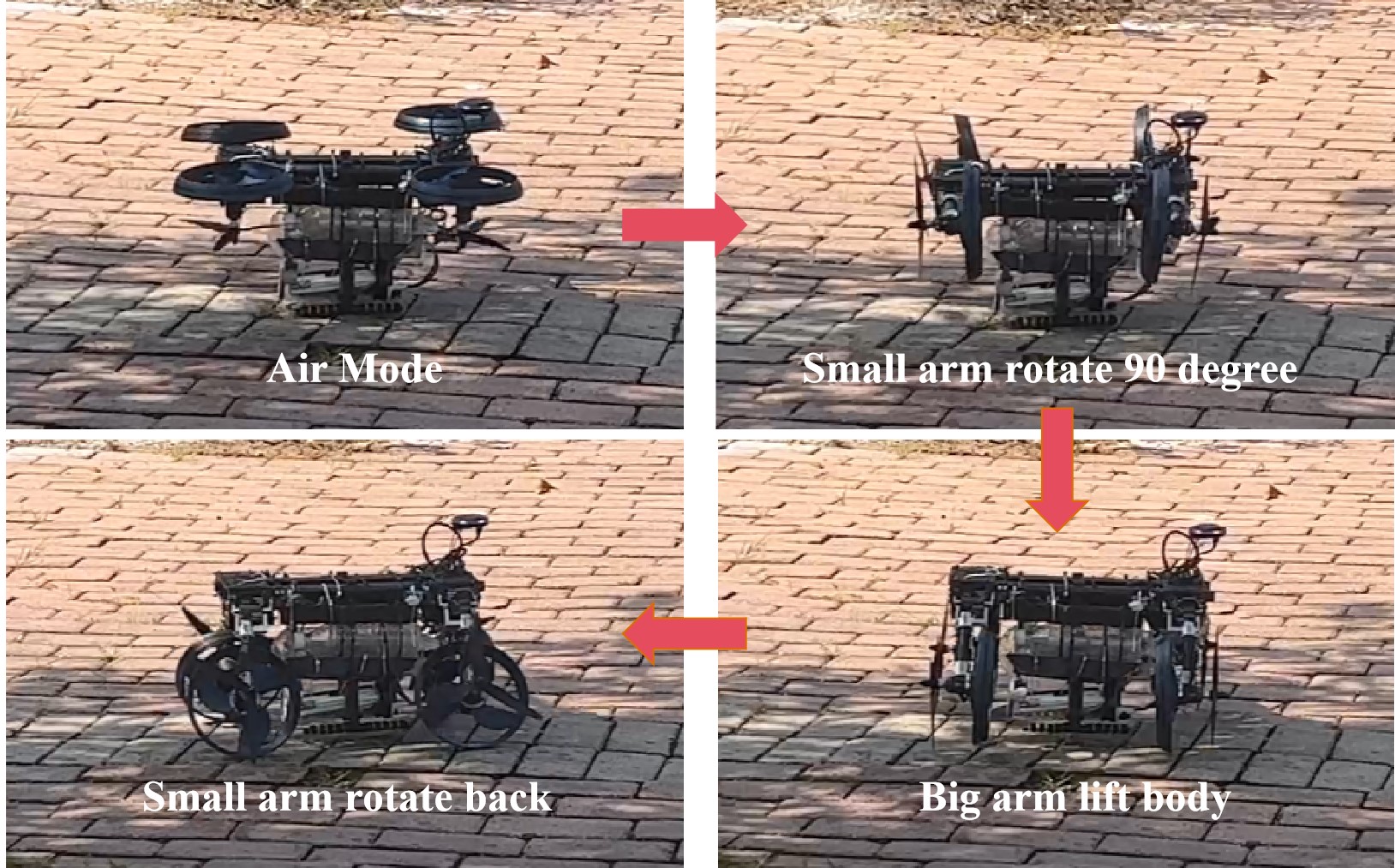}
  \caption{Air-to-Ground Mode Transition Sequence of Wukong-Omni.}
  \label{transpic}
   \vspace{-4 mm} 
\end{figure}

\subsubsection{\textbf{2) Mode Transition Strategy:}}
With the above structural design established, we develope a transition strategy tailored to the multi-domain constraints. In air-to-ground transitions, the big arm is not lifted directly, as this would require large lateral friction forces and impose excessive load on the servos and drivetrain. Instead, we exploit the natural sliding characteristics of the wheels. The small arm is first tilted so the wheels align with the sliding direction, after which the body is lifted. This approach substantially improves the reliability of transitions in uneven terrain. The full switching process is illustrated in Fig. \ref{transpic}.

As shown in Fig. \ref{wpstrapic}.B, during execution, a sinusoidal trajectory—rather than a step command—is used to drive the servos, i.e.,
\begin{align}
s_j = -&d\cos\!\left(\frac{i\pi}{N-1}\right), \\
&T_d = \frac{D_s}{N},
\end{align}
where $s_j$ denotes the servo output PWM signal, $d = +1$ indicates the lifting phase, and $d = -1$ indicates the lowering phase. The variable $i = 0, 1, \ldots, N-1$ is the step index, and $N$ is the total number of steps. Here, $T_d$ represents the time per step, and $D_s$ is the total duration. This ensures smooth transitions and allows the trajectory period to be tuned to balance switching speed, smoothness, and the signal output precision. In contrast, step commands induce servo stall currents and large instantaneous impacts, which are detrimental to the system.

With this strategy, the robot can achieve smooth and reliable transitions, significantly improving switching success rates in complex environments.

In summary, the hybrid control strategy significantly improves Wukong-Omni’s transition performance, ensuring stable, responsive, and reliable operation across different domains.

\begin{table}[!t]
  \caption{\label{tab:param}Model Parameters}
  \centering
  \resizebox{0.8\columnwidth}{!}{
  \begin{tabular}{c c}
  \hline
  Symbol&Value \\
  \hline
  $[I_{xx},I_{yy},I_{zz}]$&$[1.44,1.48,2.13]\times 10^{-1}\ \text{kg}\cdot \text{m}^2$\\
  $[m_a,V]$ & $[4.4$kg$,3.21\times10^{-3}\ \text{m}^3]$ \\
  $[L_a,W_a,H_a]$ & $[44,22,20]$\ cm\\
  $[\rho_a,\rho_w]$ & $[1.29,1000]\ \text{kg/m}^3$\\
  $\boldsymbol{K}_d,C_d$& $diag$$([0.9,0.5,0.8]),0.9$\\
  g& $9.81\ \text{m/s}^2$\\
  $[K_{TAa},K_{TAb},K_{TAc}]$& $[2.4\hspace{-1mm}\times\hspace{-1mm} 10^{-7},-4.3\hspace{-1mm}\times\hspace{-1mm} 10^{-4},-0.02]$\\
  $[K_{TMa},K_{TMb},K_{TMc}]$& $[3.91\hspace{-1mm}\times \hspace{-1mm}10^{-4},1.5\times \hspace{-1mm}10^{-7},0.216]$\\
  $[K_{TLa},K_{TLb},K_{TLc}]$& $[0,1,0]$\\
  $[K_{QA},K_{QM},K_{QL}]$& $[0.014,0.027,0]$ \\
  $[d_a,d_b,d_f,d_s]$& $[26,19,18,18]$\ cm \\
  $[\zeta,T_{L1},L_c]$& $[0.75,5,0.34]$ \\
  \hline
  \end{tabular}
  }
  \label{modelparam}
  \vspace{-4 mm} 
\end{table}

\section{6 EXPERIMENTS}
In this section, we present the experimental results to verify Wukong-Omni's effectiveness in complex, multi-domain operations and its advanced design and control capabilities. Model parameters are set as shown in TABLE \ref{modelparam}. 
\subsection{6.1 Cross-Domain Transition Capability Demonstration}
\subsubsection{\textbf{1) Water-Air Transition Capability Demonstration:}}
This section presents experimental validation of Wukong-Omni’s capability to transition between underwater and aerial modes in a rescue scenario, demonstrating its ability to overcome obstacles by rapidly switching from stable flight to submerged operation and then returning to the air.

\begin{figure*}[!t]
  \centering
  \includegraphics[width=\textwidth]{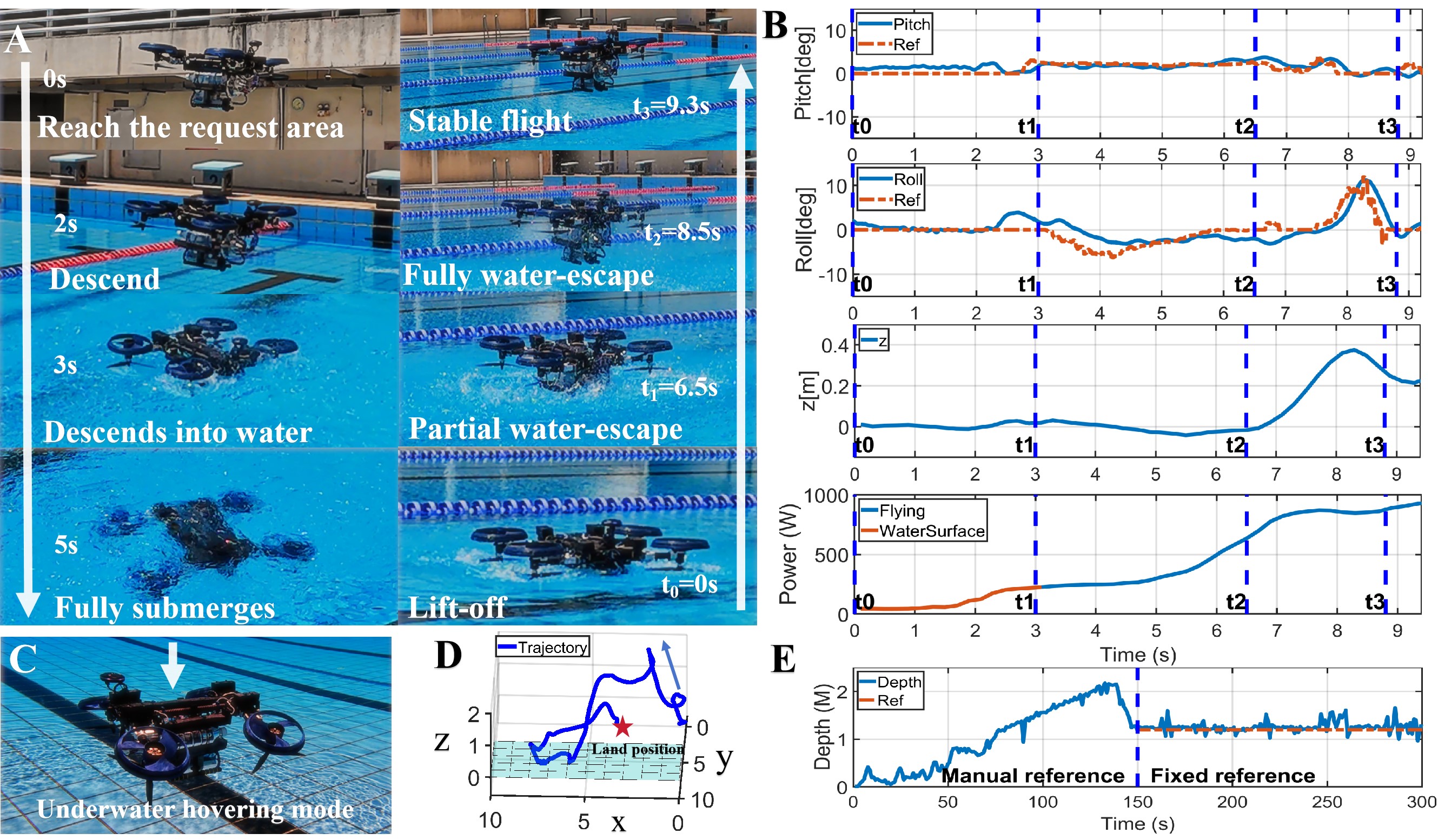}
  \caption{\textbf{A}. Complete water-air cross-domain motion loop.  
 \textbf{B}. Experimental results showing attitude and power data during the transition.  
 \textbf{C}. The robot hovers underwater after submerging.  
 \textbf{D}. Trajectory of the water-to-air transition loop.  
 \textbf{E}. Underwater depth-hold performance results.}
   \vspace{-2mm} 
   \label{exp1pic}
\end{figure*}

As illustrated in Fig. \ref{exp1pic}.A, Wukong-Omni initiates a cross-domain motion cycle by descending from air and landing on the water surface. At this point, the motors reverse to drive the wheel-propeller, initiating the dive sequence. In the shallow submersion phase, the arms are partially folded as Fig. \ref{exp1pic}.C shown, to ensure the wheel-propellers are fully submerged for effective underwater propulsion. The corresponding depth-hold data is shown in Fig. \ref{exp1pic}.E. During the initial manual descent phase, the robot rapidly dives. Once it reaches the commanded depth of approximately 1.2m, Wukong-Omni achieves relatively stable hovering. Minor fluctuations observed in the depth data are attributed to motion-induced disturbances, with a maximum deviation within 30cm, indicating robust underwater stability.

After maintaining a steady hover for a period, Wukong-Omni initiates the resurfacing maneuver at $t_0$ = 0. The four motors switch to aerial mode and activate the propellers in reverse to generate upward thrust. As control transitions to the flight controller, the aerial propellers push the robot upward and gradually break the water surface. The transition from underwater to airborne flight is completed in just 6.5 seconds and between lift-off to stable flight just 9.3 seconds, underscoring the effective design of the robot’s buoyancy, center of gravity, and thruster placement. The flight data and trajectory during the transition are presented in Fig. \ref{exp1pic}.C–D. As shown in Table \ref{EXP1table}, during the takeoff phase, the average error remains within 1.5 degrees, reflecting the structure’s strong resistance to disturbance and mechanical robustness during domain transitions.

\begin{table}[t]
 \caption{Evaluation of robot Performance in water-air Transition}
 \centering
 \resizebox{\columnwidth}{!}{
 \begin{tabular}{cccc}
 \hline
 Performance & Value& Performance & Value\\
 \hline
 Roll RMSE(deg)&1.54&Roll $\bar{e}$(deg)&1.50\\
 Pitch RMSE(deg)&1.86&Pitch $\bar{e}$(deg)&1.37\\
 Depth RMSE(m)&0.11&Depth $\bar{e}$(m)&0.07\\
 Max Surface Power(W)&220&Stable Flight Power(W)&860\\
 \hline
 \end{tabular}
 }
 \label{EXP1table}
 \vspace{-2 mm} 
\end{table}

\begin{table}[t]
\caption{Evaluation of robot Performance in land-air Transition}
\centering
 \resizebox{\columnwidth}{!}{
\begin{tabular}{cccc}
\hline
Performance & Value& Performance & Value\\
\hline
Roll RMSE(deg)&1.78&Roll $\bar{e}$(deg)&1.45\\
Pitch RMSE(deg)&2.19&Pitch $\bar{e}$(deg)&1.77\\
Yaw RMSE(deg)&7.40&Yaw $\bar{e}$(deg)&4.54\\
Max Land Power(W)&74.2&Max Flight Power(W)&1010\\
\hline
\end{tabular}}
 \label{exp2table}
 \vspace{-5 mm} 
\end{table}

\subsubsection{\textbf{2) Ground-Air Transition Capability Demonstration:}}
Following the verification of air-water transitions, this section evaluates Wukong-Omni’s ability to switch modes when encountering tall ground obstacles or accessing elevated structures in rescue scenarios. The validation is conducted through multiple full transition cycles in an outdoor environment.

\begin{figure*}[!t]
  \centering
  \includegraphics[width=\textwidth]{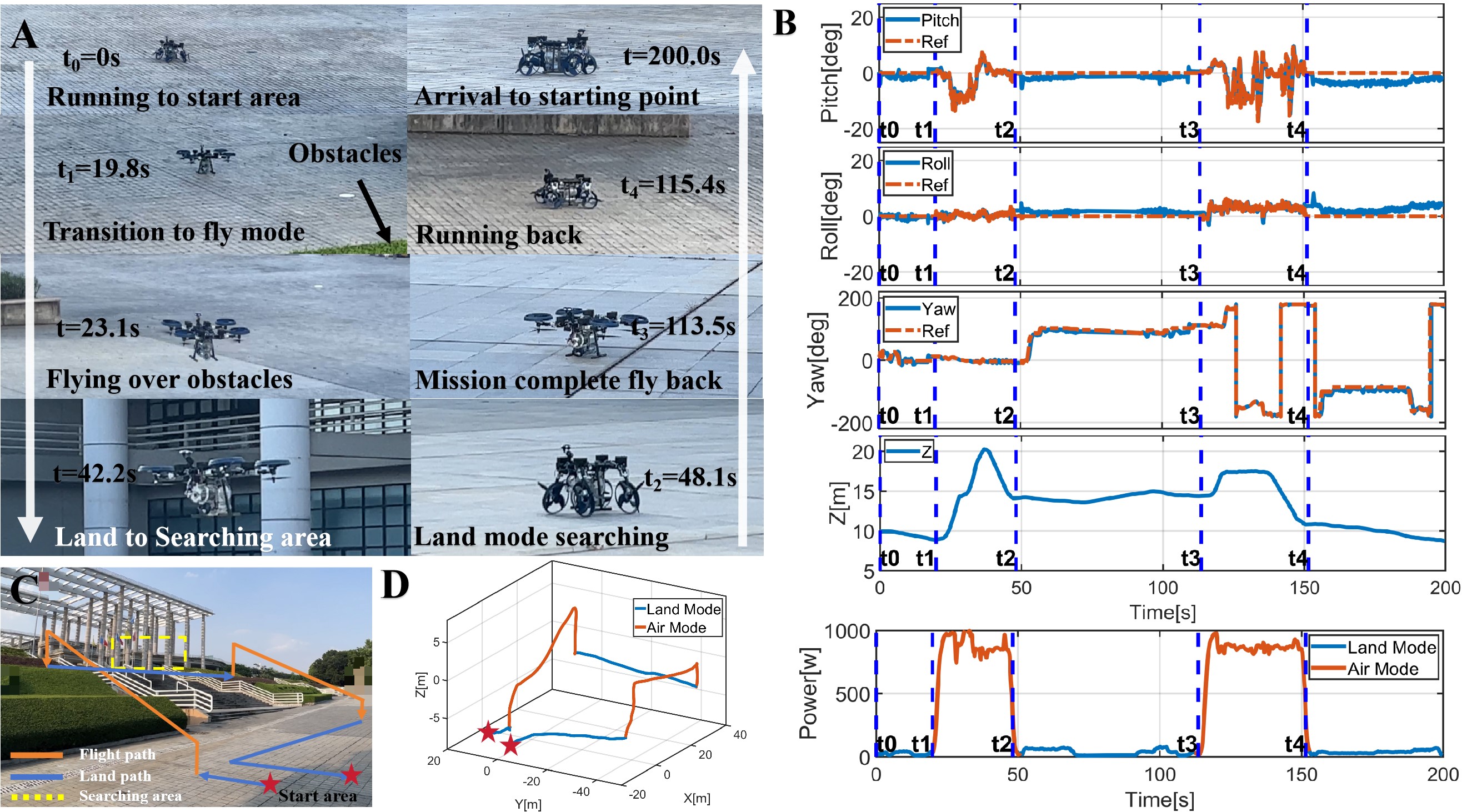}
\caption{\textbf{A}. Complete air-land cross-domain motion loop.  
 \textbf{B}. Experimental results showing attitude and power data during the transition.  
 \textbf{C}. The robot mission path in field test.  
 \textbf{D}. Trajectory of the air-land transition loop.}
    \vspace{-2 mm} 
\label{exp2pic}
\end{figure*}

As shown in Fig.\ref{exp2pic}.A, Wukong-Omni begins exploring in ground mode at $t_0$ = 0. At $t_1$ = 19.8, it arrives at the base of the obstacle and switches to aerial mode to take off. By $t_2$ = 42.2, it has successfully flown over the obstacle and landed in the target search area, where it transitions back to ground mode for further terrestrial exploration. After navigating the area, it takes off again at $t_3$ = 113.5 and lands on the nearest surface at $t_4$ = 115.4, finally returning to the starting point by $t$ = 200 to complete the entire loop.

As illustrated in Fig. \ref{exp2pic}.B, the land-air transition is accomplished in just 6 seconds, showcasing Wukong-Omni’s suitability for rapid cross-domain switching in time-critical rescue missions. The cross-domain trajectory and schematic are shown in Fig. \ref{exp2pic}.C–D. During obstacle traversal, the robot ascends 10 m within seconds to overcome the barrier. After landing, it switches back to ground mode and continues exploration over a 40 m path with minimal energy consumption, which traditional search robots are typically unable to achieve. As shown in Table \ref{exp2table}, the altitude RMSE, mean error, and maximum power highlight Wukong-Omni’s exceptional cross-domain mobility and high-efficiency task performance.

\begin{figure*}[!t]
  \centering
  \includegraphics[width=\textwidth]{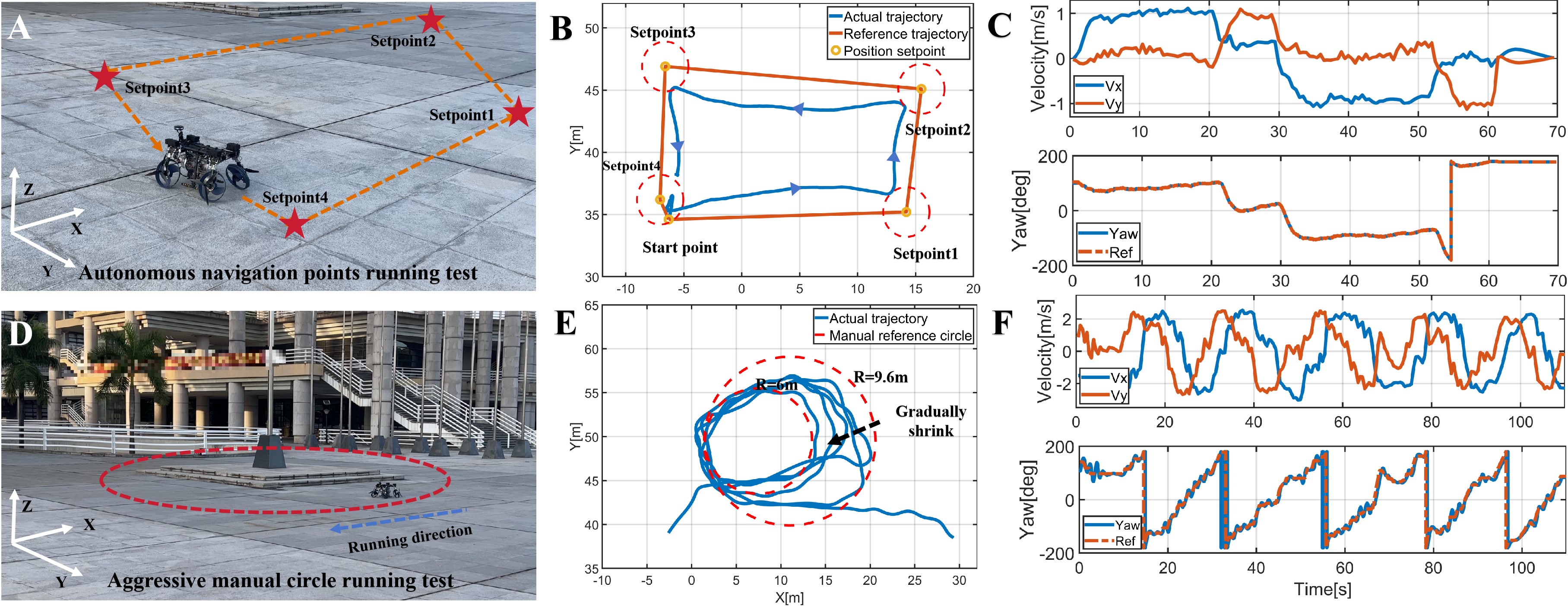}
\caption{
    \textbf{A}. Wukong-Omni tracking a designated path in land mode.      
    \textbf{B}. Wukong-Omni performing manual circling in land mode.  
    \textbf{C}. Experimental results showing the trajectory, velocity, and yaw during tracking.  
    \textbf{D}. Experimental results showing the trajectory, velocity, and yaw during the circling test.}
  \label{exp4pic}
   \vspace{-4 mm} 
\end{figure*}

\subsection{6.2 Each Mode Performance Assessment Experiments}
\subsubsection{\textbf{1) Land Trajectory Tracking Test:}}
In the ground rescue scenario, Wukong-Omni utilizes the L1 controller to autonomously navigate between waypoints when GPS signals are available. We designed an irregular quadrilateral route with a length of 20 meters and a width of 10 meters, as shown in Fig. \ref{exp4pic}.A. The actual test environment is shown in Fig. \ref{exp4pic}.B and Fig. \ref{exp4pic}.C, where the trajectory demonstrates smooth circular transitions between waypoints and a cruising speed of 1.1 m/s. The yaw axis responds rapidly. Additionally, we conducted a manually controlled aggressive driving test, where the vehicle continuously reduced its turning radius around a flagpole. As shown in Fig. \ref{exp4pic}.D, the turning radius decreased from 9.6 meters to 6 meters, while the maximum speed was still maintained at 2.1 m/s. The experimental data in Fig. \ref{exp4pic}.E and Fig. \ref{exp4pic}.F show that even at relatively high speeds, the yaw axis maintained good tracking performance, with only minor angular fluctuations, the indicators can be found in Table \ref{exp3table}. The two experiments demonstrate that Wukong-Omni possesses strong maneuverability and a certain degree of autonomy in ground mode.

\subsubsection{\textbf{2) Air Trajectory Tracking Test:}}

\begin{table}[t]
\caption{Evaluation of Robot Performance In Land Mode}
\centering
 \resizebox{\columnwidth}{!}{
\begin{tabular}{cccc}
\hline
Cruise Performance & Value& Manual Performance &Value \\
\hline
Yaw $\bar{e}$(deg)&0.08&Yaw $\bar{e}$(deg)&8.71\\
Yaw RMSE(deg)&0.53&Yaw RMSE(deg)&10.5\\
Speed(m/s)&1.0&Max Speed(m/s)&3.3\\
Stop Radius(m)&2.5&Max Power(W)&152\\
Average Power(W)&27.36&Average Power(W)&76\\
\hline
\end{tabular}}
 \label{exp3table}
 \vspace{-6 mm} 
\end{table}

\begin{figure*}[!t]
  \centering
  \includegraphics[width=\textwidth]{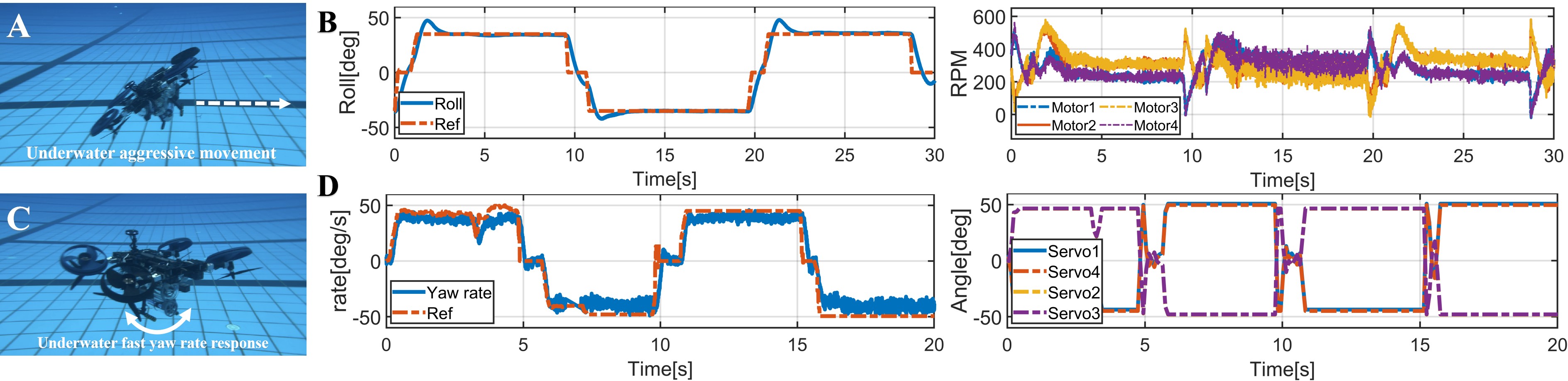}
\caption{
\textbf{A}. Schematic illustration of underwater lateral motion.  
\textbf{B}. Underwater roll attitude response and corresponding motor speed.  
\textbf{C}. Schematic illustration of underwater yaw maneuver.  
\textbf{D}. Underwater yaw rate response and tilt servo actuation data.
}
\label{exp6pic}
\vspace{-3 mm} 
\end{figure*}

\begin{figure}[!t]
  \centering
  \includegraphics[width=\columnwidth]{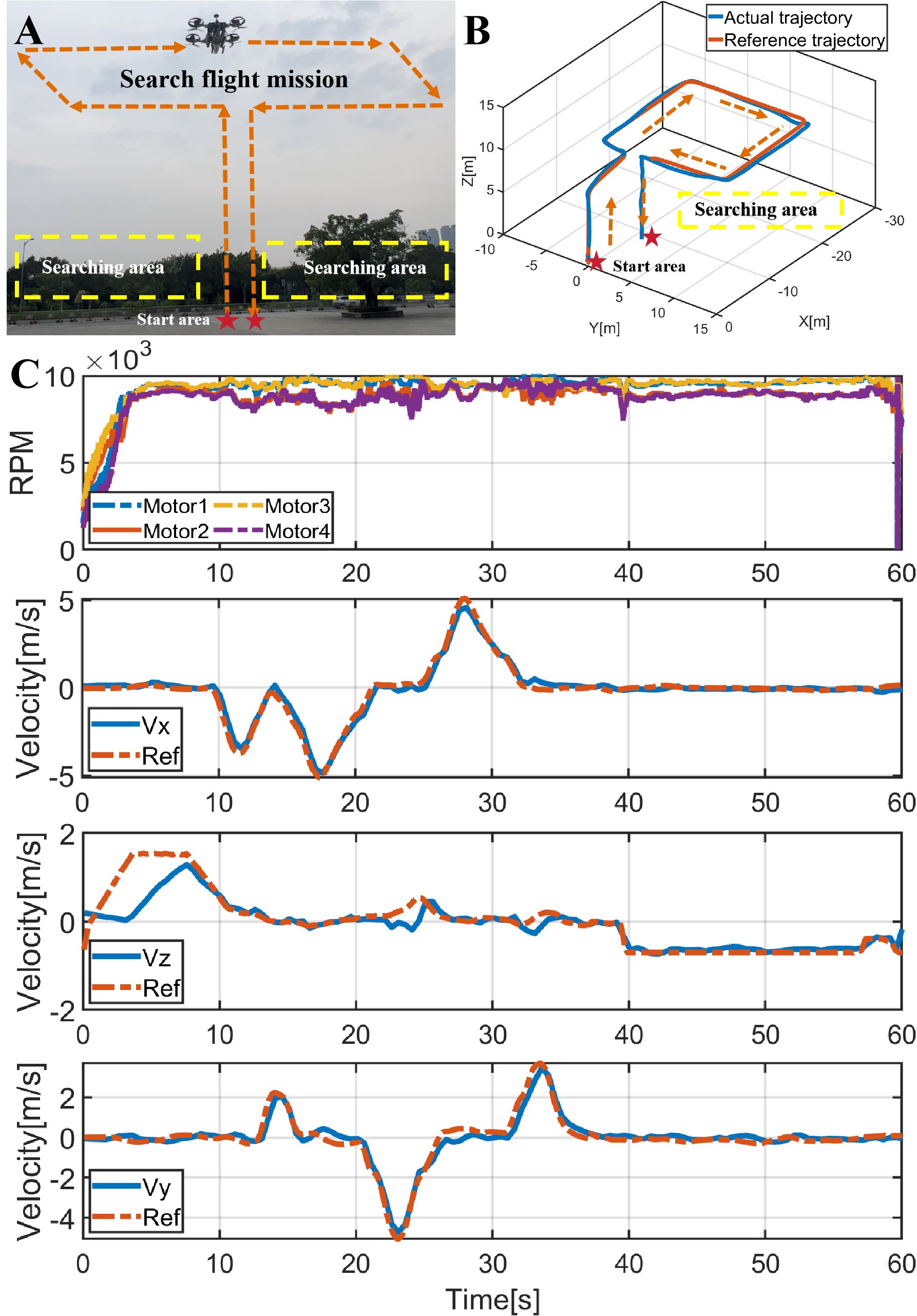}
\caption{
\textbf{A}. Schematic illustration of trajectory tracking in aerial mode. 
\textbf{B}. Experimental results of trajectory tracking. 
\textbf{C}. Experimental results of body velocity and motor rotational speed.
}
\label{exp5pic}
\vspace{-4 mm} 
\end{figure}

In addition to obstacle crossing, the aerial mode of Wukong-Omni plays a crucial role in rapid aerial search during rescue missions. When GPS signals are available, the vehicle is capable of executing predefined waypoint plans to autonomously perform aerial tasks. As shown in Fig. \ref{exp5pic}.A, it surveys the yellow search area from the air to detect high-value targets. We designed a 16m×16m waypoint trajectory for the robot to autonomously depart from its initial position, navigate the waypoint area, upon task completion. The trajectory is illustrated in Fig. \ref{exp5pic}.B. Flying at an altitude of 10 meters, Wukong-Omni closely follows the planned path and achieves a maximum speed of 5 m/s, enabling fast and efficient flight. Flight data are shown in Fig. \ref{exp5pic}.C. As shown in Table \ref{exp3table} these results demonstrate Wukong-Omni’s outstanding aerial performance and confirm that the all-in-one propulsion unit enables stable and reliable execution of complex missions in the air.
\subsubsection{\textbf{3) Underwater Altitude Tracking Test:}}

In this section, we demonstrate its high maneuverability underwater through attitude response experiments. Since the control methods for the roll and pitch axes are similar, only the roll and yaw axis data are presented. Underwater operation is shown in Fig. \ref{exp6pic}.AC, and the corresponding experimental results are illustrated in Fig.\ref{exp6pic}.BD. The roll axis is capable of controlled large-angle maneuvers up to 35 degrees. Benefiting from the tiltable yaw axis design, the yaw angular velocity can still reach 45 degrees per second underwater, with the thruster speed stabilized at approximately 350 RPM and peaking at 590 RPM during rapid maneuvers. All values remain within the stable operating range of the all-in-one propulsion unit, the results as shown in Table \ref{exp4table} confirming its excellent performance and adaptability in complex underwater scenarios.

\begin{figure}[t]
  \centering
  \includegraphics[width=\columnwidth]{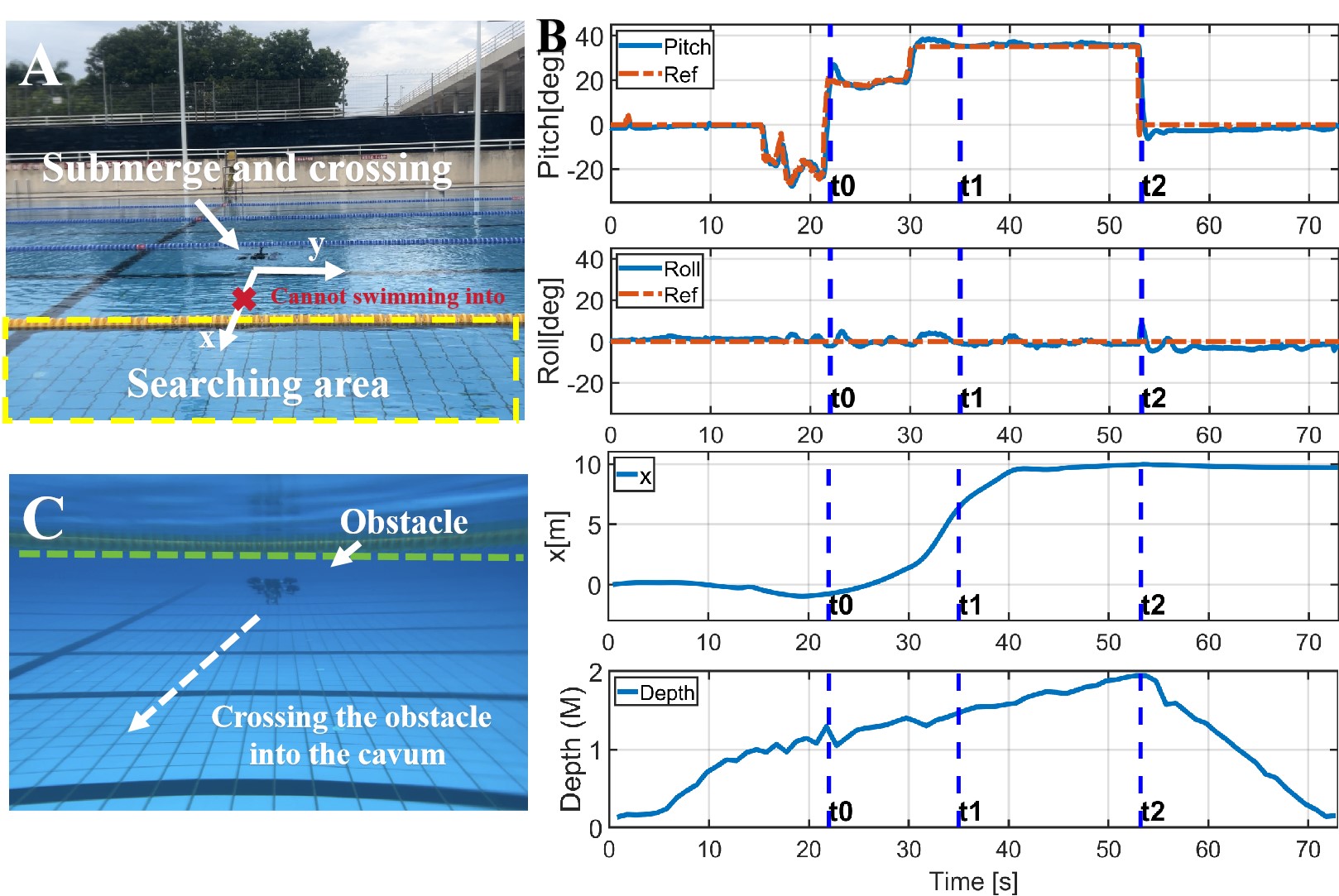}
  \caption{ 
  \textbf{A}. The Obstacles on the water surface and the coordinate system which they are traversed. 
  \textbf{B}. Experimental results of the attitude and position during the traversed.
  \textbf{C}. Underwater crossing process pictures.}
  \label{exp3pic}
   \vspace{-2mm} 
\end{figure}

\begin{table}[t]
\caption{Evaluation of Robot Performance In Air And Water Mode}
\centering
 \resizebox{\columnwidth}{!}{
\begin{tabular}{cccc}
\hline
Air Performance&value&Water Performance&value\\
\hline
Trajectory $\bar{e}$(m)&0.31&Roll $\bar{e}$(deg)&3.33\\
Trajectory RMSE(m)&0.35&Roll RMSE(deg)&6.56\\
Max Speed(m/s)&4.86&Max Speed(m/s)&1.2\\
Average Power(W)&942&Average Power(W)&128\\
\hline
\end{tabular}}
 \label{exp4table}
 \vspace{-5 mm} 
\end{table}

\subsection{6.3 Underwater Entry via Surface Obstacle perfomance assessment}
Most existing cross-domain robots are limited to floating or surface navigation, restricting their use in flooded environments. In contrast, Wukong-Omni is capable of diving and traversing water-surface obstacles, allowing access to submerged structures during search-and-rescue missions. To validate this, we conducted an obstacle traversal experiment using yellow floating barriers to simulate the walls of a flooded building (Fig. \ref{exp3pic}.A). The robot, manually controlled due to GPS unavailability underwater, hovered near the barrier, dived to a specified depth, advanced underwater, surfaced, and entered the marked search zone (Fig. \ref{exp3pic}.C). Wukong-Omni reached a depth of 2m and pitched up to 35 degrees to accelerate forward, covering 10m underwater (Fig. \ref{exp3pic}.B). This demonstration highlights Wukong-Omni’s underwater mobility, which is an essential capability lacking in most cross-domain robots.
\subsection{6.4 Field Mission Assessment}
\begin{figure}[t]
  \centering
  \includegraphics[width=\columnwidth]{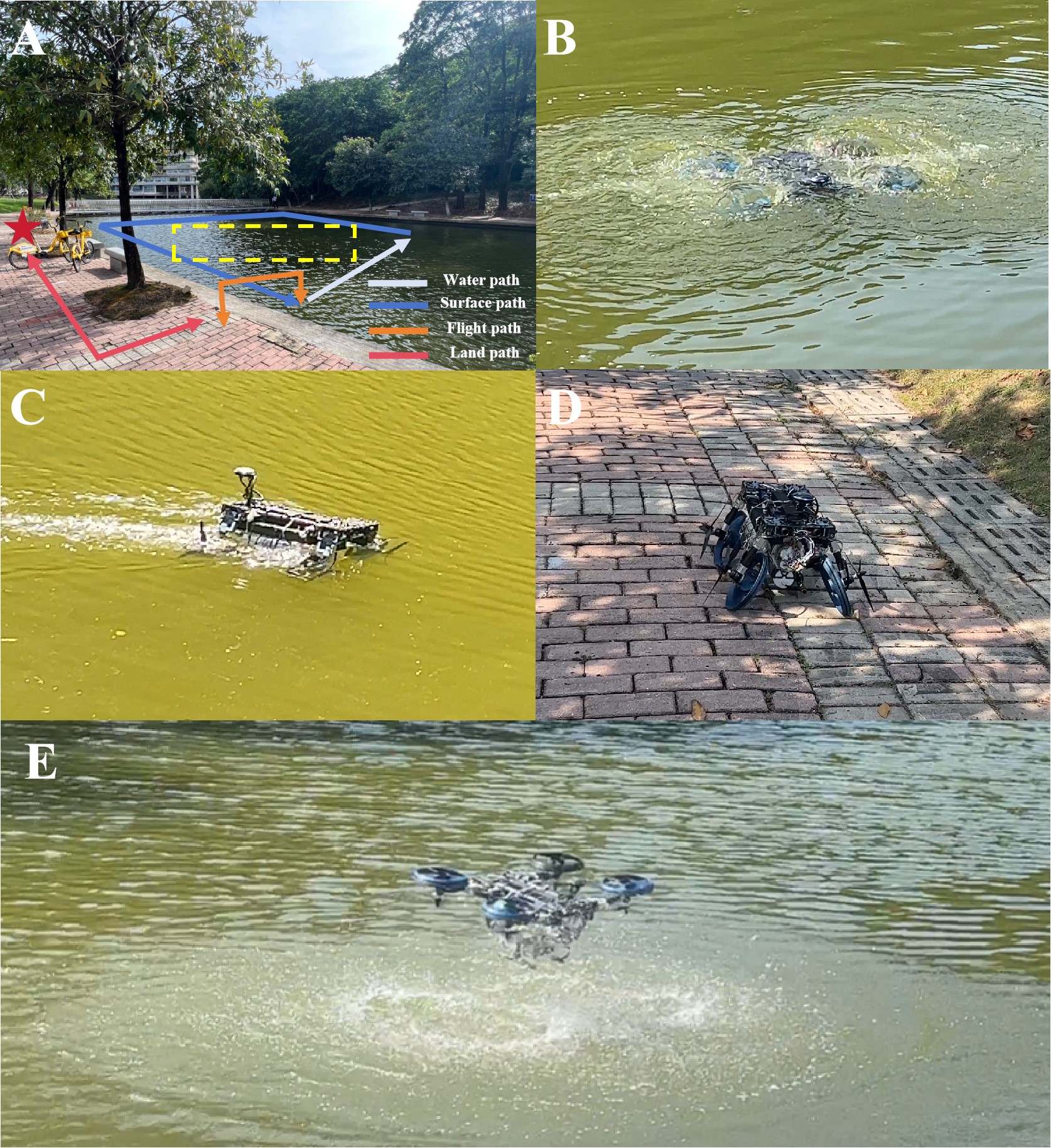}
  \caption{ 
  \textbf{A}. Illustration of the field mission execution.
\textbf{B–E}. Wukong-Omni successfully performing outdoor tasks at an artificial lake, including water surface moving, underwater operation, ground locomotion, and air–water transitions.}
  \label{fieldpic}
   \vspace{-2 mm} 
\end{figure}

Wukong-Omni was comprehensively tested at an outdoor environment lake. As shown in Fig. \ref{fieldpic}.A, the robot executed a full inspection-and-return mission. It first traveled to the lakeside in land mode Fig. \ref{fieldpic}.D, and then transitioned to aerial mode to take off and land on the water surface. After landing, it switched to underwater mode and dived to simulate an underwater inspection, moving toward the center of the lake Fig. \ref{fieldpic}.B. It then resurfaced and performed an on-surface task, with a rectangular trajectory on the water Fig. \ref{fieldpic}.C. Finally, it returned to the starting point, performed a water–air transition (Fig. \ref{fieldpic}.E), switched back to land mode, and completed the mission.
This experiment demonstrates that Wukong-Omni can operate reliably across all modes in real outdoor environments.

\begin{figure}[t]
  \centering
  \includegraphics[width=\columnwidth]{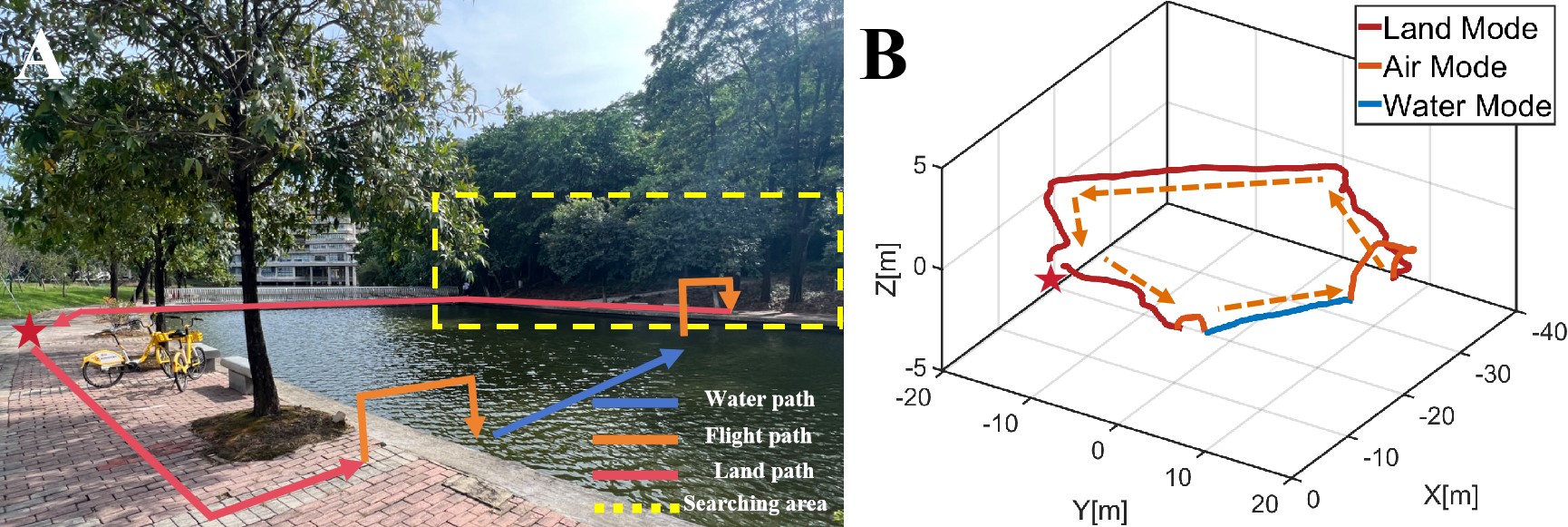}
  \caption{ 
  \textbf{A}. Illustration of the mission execution. \textbf{B}. Outdoor experiment trajectory.}
  \label{fieldcycpic}
   \vspace{-5mm} 
\end{figure}

We further increased the mission difficulty. Wukong-Omni was commanded to leave the starting point, to perform multiple water–air transitions and on-surface navigation, and land on the opposite shore of the lake—simulating a long-range operational task Fig. \ref{fieldcycpic}.A. It then returned to the starting point by crossing a bridge using land mode, the trajectory as shown in Fig. \ref{fieldcycpic}.B. Without battery replacement or maintenance, the robot completed this challenging mission four consecutive times, achieving a 100\% success rate. Across all trials, Wukong-Omni successfully executed multiple cross-domain transitions and navigated uneven terrains, demonstrating excellent robustness and outdoor operational reliability suitable for demanding tasks such as search and rescue.
\section{7 CONCLUSION}
This study presents a novel three-domain robot capable of operating across land, water, and air using all-in-one propulsion unit. By leveraging the deformable design of the robot's main body, the system eliminates the need for multiple actuators traditionally required for cross-domain operation. This not only supports a more compact mechanical structure but also reduces additional energy consumption caused by redundant propulsion systems.

The all-in-one propulsion unit combines both propellers and wheels, whose mutual interactions and geometric parameters critically influence the robot’s performance across different domain. To address this, we propose a design and optimization methodology tailored for three-domain mobility, integrating theoretical modeling, simulation, and experimental validation to optimize the all-in-one propulsion unit.

A prototype of the three-domain robot, Wukong-Omni, was fabricated and tested in both pool and outdoor environments to evaluate its performance across domains. Experimental results demonstrate the robot’s ability to seamlessly transition among the three domain and execute a range of tasks. In aerial mode, Wukong-Omni autonomously follows waypoint trajectories at altitudes of 10 meters with a top speed of 5 m/s. On land, it can navigate autonomously using the L1 control algorithm or be manually operated with maximum running speed of 2.1 m/s. In aquatic trials, the robot exhibited stable underwater hovering, agile roll maneuvers up to 35 degrees, and yaw rotation speeds sustained at 45 deg/s.

These results highlight Wukong-Omni’s exceptional locomotion capabilities, including precise attitude and depth control, as well as autonomous path-following across diverse environments. This makes Wukong-Omni a highly promising platform for efficient and adaptable operations in flood-related search and rescue missions.

\begin{acks}
The authors would like to thank Keqi Luo from the School of Automation, Guangdong University of Technology for his valuable input on refining the equations and improving the manuscript, and Tao Yang from Guangzhou XAG Co.Ltd for his assistance in designing the all-in-one unit and contributing to the development of the performance metrics.
\end{acks}
\begin{dci}
The author(s) declared no potential conflicts of interest with respect to the research, authorship, and/or publication of this article.
\end{dci}
\begin{funding}
The authors disclosed receipt of the following financial support for the research, authorship, and/or publication of this article:This work was supported by the National Natural Science Foundation of China under Grant (U21A20476, 62121004), Guangdong Innovative and Research Teams Project under Grant (2019ZT08X340 and 2019BT02X353), and grants (AoE/E-601/24-N, 16203223 and C602923G) from the Research Grants Council of the Hong Kong Special Administrative Region, China.
\end{funding}
\section{ORCID IDs}
Yufan Liu ~\orcidlink{0000-0002-7144-7665}\href{https://orcid.org/0000-0002-7144-7665}{https://orcid.org/0000-0002-7144-7665}\\
Rixi Yu ~\orcidlink{0009-0001-5311-7806}\href{https://orcid.org/0009-0001-5311-7806}{https://orcid.org/0009-0001-5311-7806}\\
Junjie Li ~\orcidlink{0009-0005-4722-0240}\href{https://orcid.org/0009-0005-4722-0240}{https://orcid.org/0009-0005-4722-0240}\\
Wei Meng ~\orcidlink{0000-0002-8513-5013}\href{https://orcid.org/0000-0002-8513-5013}{https://orcid.org/0000-0002-8513-5013}\\
Fumin Zhang ~\orcidlink{0000-0003-0053-4224}\href{https://orcid.org/0000-0003-0053-4224}{https://orcid.org/0000-0003-0053-4224}
\bibliographystyle{SageH}
\bibliography{reference}

\end{document}